\documentclass[acmsmall,nonacm]{acmart}

\usepackage{array}
\usepackage{booktabs}
\usepackage{longtable}
\usepackage{tikz}
\usetikzlibrary{positioning,fit,arrows.meta,shapes.geometric,shapes.symbols}
\usepackage{xurl}
\usepackage{newunicodechar}

\newunicodechar{’}{\textquotesingle}
\newcommand{\cid}[1]{\texttt{\nolinkurl{#1}}}
\hypersetup{
  hidelinks,
  pdftitle={A Deterministic and Auditable AI Security Risk Assessment Framework with ATLAS Aligned Executable Rules and Formal Verification},
  pdfauthor={Yixuan Huang, Basel Halak, Boojoong Kang},
  pdfdisplaydoctitle=true
}

\AtBeginDocument{%
  }

\begin{document}

\title{A Deterministic and Auditable AI Security Risk Assessment Framework with ATLAS Aligned Executable Rules and Formal Verification}

\author{Yixuan Huang}
\affiliation{%
  \institution{University of Southampton}
  \city{Southampton}
  \state{England}
  \country{UK}
}
\email{yh3c24@soton.ac.uk}

\author{Basel Halak}
\affiliation{%
  \institution{University of Southampton}
  \city{Southampton}
  \state{England}
  \country{UK}
}
\email{Basel.Halak@soton.ac.uk}

\author{Boojoong Kang}
\affiliation{%
  \institution{University of Southampton}
  \city{Southampton}
  \state{England}
  \country{UK}
}
\email{B.Kang@soton.ac.uk}

\begin{abstract}
Artificial intelligence systems are increasingly deployed in high impact and safety critical settings, yet security assessment remains difficult to reproduce and defend under audit. Existing approaches often rely on narrative checklists or assessor driven scoring, and they lack an explicit, machine evaluable mapping from observable engineering artefacts to stable technique level outcomes. We present an evidence driven AI security assessment framework that operationalises assessment as a deterministic decision function. The framework normalises heterogeneous artefacts into a project independent Control ID taxonomy scored on a bounded four level ordinal scale, compiles technique level predicates from a pinned MITRE ATLAS snapshot via an explicit mitigation to control mapping, and outputs technique indexed feasibility and impact levels with traceable links back to the triggering evidence. We package all normative choices as a versioned assessment policy object to support repeatable reassessment across snapshots. To ensure semantic correctness, we formally verify boundedness, totality, ordered semantic consistency, and monotonicity of the compiled evaluator over the full declared score domain. We evaluate the framework on five public open source AI projects pinned to explicit repository snapshots, quantify before and after changes under a unified hardening intervention, and validate responsiveness to real engineering changes through fork based implementations of Software Bill of Materials (SBOM) generation and Continuous integration (CI) security scanning gates. Results show consistent downward shifts in feasibility profiles under strengthened observable controls, while worst case residual feasibility persists when technique specific core controls remain absent from the evidence scope.
\end{abstract}


\maketitle

\section{Introduction}

Artificial intelligence systems are increasingly deployed in high impact and safety critical environments, including financial services, healthcare, and autonomous infrastructure\cite{intro}. Evaluating these deployments requires rigorous risk assessment, which typically encompasses both safety and security. Safety concerns unintended failures that arise from model errors, distributional shift, or system malfunction. Security, on the other hand, addresses intentional adversarial actions that exploit system interfaces, supply chains, data pipelines, or operational processes to compromise integrity, confidentiality, or availability. While safety and security may interact, they represent distinct risk domains and require different assessment semantics\cite{safety}. Despite the critical need to evaluate these intentional threats, methodologies for AI security risk assessment remain immature and difficult to reproduce under audit. Many existing methodologies rely on qualitative checklists, narrative justifications, or assessor driven scoring matrices. Such methods lack a deterministic mapping from observable engineering evidence to concrete risk outputs. As a result, different assessors may interpret the same artefacts differently, evidence is not consistently normalised, and assessment outcomes are difficult to justify when procurement decisions, release gates, or regulatory reviews require a traceable chain from artefact to conclusion.

The fundamental gap in current AI security risk assessments is the absence of an executable and auditable decision function capable of reliably mapping verifiable artefacts to well defined risk indicators. Because existing frameworks lack this deterministic mapping, they fail to satisfy three critical requirements for reproducible assessment. First, rather than grounding inputs in inspectable engineering artefacts, they often rely on subjective narrative statements. Second, they struggle to normalise heterogeneous evidence into a stable abstraction, which prevents reliable comparison across different projects. Third, their evaluation semantics are typically implicit or assessor-dependent, meaning that assessment outputs cannot be consistently reproduced under a fixed policy and system snapshot.

To address this lack of deterministic mapping and fulfill these three requirements, we introduce an evidence driven AI security risk assessment framework. First, to ensure inputs are grounded in inspectable evidence, the framework directly ingests raw engineering artefacts, such as dependency manifests, vulnerability scanning outputs, provenance metadata, and deployment configurations. Second, to achieve stable normalisation, it transforms these heterogeneous artefacts into a project-independent Control ID taxonomy, where Control IDs serve as atomic control variables that map evidence to bounded ordinal scores. Third, to provide explicit and reproducible evaluation semantics, technique level decision logic is compiled into machine-executable predicates derived from MITRE ATLAS technique mitigation relations. By evaluating these predicates under explicit ordered semantics, the framework produces discrete feasibility and impact levels for each technique, providing a fully traceable link from the final output back to the underlying engineering artefacts.

A distinguishing mechanism of our approach is that the assessment mechanism is formally specified and verified. Because the framework evaluates bounded ordinal control evidence through compiled predicates, correctness depends not only on control coverage but also on the logical soundness of the decision semantics. We therefore introduce a formal verification layer that proves boundedness, totality, ordered semantic consistency, and monotonicity under strengthened controls over the entire declared input domain. These guarantees establish that the evaluator is a well defined and governance ready decision function.

We empirically evaluate the framework on five public open source AI projects under pinned repository snapshots. To probe responsiveness under a controlled perturbation, we apply a unified hardening intervention across projects: we strengthen the same intervention subset of observable controls by setting the corresponding Control IDs to the strongest rubric level (3) while leaving all other observed controls unchanged, and we rerun the evaluator on the same pinned snapshots under a fixed and versioned policy configuration with constant impact assignment. The results show consistent reductions in aggregate feasibility impact summaries together with clear downward shifts in technique level feasibility distributions. For example, the Feasibility Impact Summary Index decreases from 198 to 184 in KServe and from 324 to 266 in Cassava Example, and the count of techniques in the lowest feasibility band F equals 0 increases markedly, with KServe moving from 2 to 9 and vLLM moving from 0 to 9. At the same time, the worst case technique risk remains unchanged under partial strengthening, reflecting that hardening a bounded subset of controls cannot eliminate techniques whose feasibility is dominated by missing controls outside the intervention set. Taken together, these findings show that evidence driven assessment can be reproducible, auditable, and sensitive to concrete control improvements while remaining explicit about what has not been strengthened.

The scope of this work is deliberately non probabilistic. The framework does not estimate attack frequency, does not predict incident rates, and does not claim compliance with external certification regimes. Instead, it provides evidence derived feasibility indicators and ordinal impact levels computed under a fixed and versioned policy configuration. Its purpose is to support audit, cross project comparison, and change tracking across system states, rather than probabilistic forecasting.

This paper makes four contributions:
\begin{enumerate}
\item \textbf{Formally specified executable assessment semantics.}
We define an explicit technique indexed decision semantics that maps bounded ordinal control evidence to bounded feasibility and impact outputs under ordered predicate evaluation, producing an auditable evidence to output trace.

\item \textbf{A versioned and auditable policy configuration model.}
We introduce a policy object $\mathcal{P}$ that packages normative configuration choices, including the pinned ATLAS snapshot, mitigation relation, Control ID taxonomy and scoring rubrics, mitigation to control mapping $\Phi$, technique profiles $(K_t,O_t)$, and impact parametrisation, enabling reproducible reassessment and governance controlled evolution.

\item \textbf{A formally verified evaluator with bounded guarantees.}
We implement a verifier backed evaluator and discharge bounded domain proof obligations establishing boundedness, totality, ordered semantic consistency, and monotonicity, ensuring that the compiled rule base behaves correctly over the declared ordinal input domain.

\item \textbf{Empirical evaluation on pinned real world project snapshots.}
We evaluate the framework on frozen snapshots of five public open source AI projects and measure before and after changes under both counterfactual and concrete control strengthening interventions.
\end{enumerate}

The remainder of the paper is organised as follows. Section~2 reviews representative governance frameworks, threat knowledge bases, and academic approaches to motivate the operational gap addressed here. Section~3 presents the framework design and rule compilation process. Section~4 describes the formal verification. Sections~5 and~6 report the experimental evaluation and results.

\section{Background and Motivation}

This section reviews representative governance frameworks, threat knowledge bases, and academic risk models to clarify the operational gap that motivates our work. Across these strands of literature, important advances have been made in organising concepts and structuring discussions of AI risk. However, a common limitation remains: the absence of an evidence grounded, executable, and auditable decision semantics that deterministically maps observable control evidence to stable risk outputs.

Policy and industry frameworks such as NIST AI RMF 1.0 \cite{ref1} and the NCSC Machine Learning Security Principles \cite{ref2} centre AI risk management around governance loops and lifecycle guidance. They articulate control themes spanning access control, supply chain assurance, monitoring, documentation, and operational readiness. These frameworks are valuable for setting expectations and guiding oversight, yet they are primarily normative. They do not define a machine evaluable mechanism that converts verifiable artefacts into discrete and reproducible security outcomes. 

The CSA AI Model Risk Management Framework \cite{ref4} further strengthens documentation practices through model cards, risk sheets, and structured scenario analysis. Its emphasis remains procedural and descriptive. Microsoft’s AI Security Risk Assessment \cite{ref5} introduces semi quantitative scoring matrices, but the scoring logic is fixed and not formally specified as an executable rule system. As a consequence, assessments frequently depend on assessor interpretation, and the chain from artefact to conclusion is not guaranteed to be reproducible across teams or over time.

Threat knowledge bases such as MITRE ATLAS \cite{ref3} provide a structured vocabulary of adversarial tactics, techniques, and mitigations for AI enabled systems. ATLAS supports systematic coverage of attacker behaviours and improves conceptual alignment between threats and defences. However, it is not an assessment pipeline. It does not prescribe how organisational artefacts should be normalised into atomic control variables, nor how mitigation relations should be compiled into executable decision logic. The practical challenge is therefore not only identifying relevant techniques, but operationalising them into a deterministic and evidence traceable evaluator.

Academic approaches introduce additional formal structure but still leave gaps at the level of operational semantics. Bitton et al. \cite{ref6} combine Analytic Hierarchy Process weighting with MulVAL attack graphs to reason across lifecycle stages. Although expressive, the method relies on subjective weight assignment, which reduces reproducibility and auditability across organisations. Sector specific schemes such as the Exposure and Severity formulation for autonomous vehicles by Grosse and Alahi \cite{ref7} provide domain insights but lack general applicability across deployment contexts. Yajima et al. \cite{ref8} encode feasibility as Boolean conditions, obscuring graded severity and limiting prioritisation. Formal robustness analyses such as those by Xu et al. \cite{ref9} establish mathematical guarantees for local input perturbations, yet they do not address system level threats including model extraction, misuse, and supply chain compromise. Earlier risk management formulations based on multi criteria decision analysis over confidentiality, integrity, and availability similarly depend on expert assigned weights and lack a binding from concrete artefacts to deterministic severity levels.

Table~\ref{tab:rw-compare} summarises these approaches along eight dimensions relevant to operational AI security assessment. Evidence grounding captures whether inputs are tied to verifiable engineering artefacts. Control abstraction layer indicates whether heterogeneous artefacts are normalised into stable atomic variables. Executable semantics denotes whether decision logic is machine evaluable rather than narrative or fixed matrix based. Output format reflects whether results are discrete and comparable. Attack specificity records whether different adversarial techniques are assessed independently. Traceability indicates whether outputs can be linked back to triggering evidence. System level coverage evaluates whether interfaces, deployment workflows, and supply chain exposures are considered. Assessor dependence measures sensitivity to subjective interpretation. 

Across these dimensions, existing approaches improve conceptual organisation and, in some cases, introduce quantification. Yet they do not provide a deterministic and formally specified mapping from observable control evidence to technique indexed security outcomes. This operational gap motivates the development of an evidence driven framework in which governance intent is compiled into executable rules over Control IDs, producing auditable and reproducible feasibility and impact levels.

\begin{table}
  \caption{Comparison of AI Security Assessment Approaches and the Proposed Framework (KB: knowledge base; MCDA: multi criteria decision analysis; F/I: feasibility/impact).}
  \label{tab:rw-compare}
  \centering
  \small
  \setlength{\tabcolsep}{3pt}
  \renewcommand{\arraystretch}{1.05}
  \resizebox{\textwidth}{!}{%
  \begin{tabular}{p{2.6cm}cccccccc}
    \toprule
    Approach &
    \shortstack{Evidence\\Grounding} &
    \shortstack{Control\\Layer} &
    \shortstack{Executable\\Semantics} &
    Output &
    \shortstack{Attack\\Specific} &
    Traceability &
    \shortstack{System\\Level} &
    \shortstack{Assessor\\Dependence}\\

    \midrule
    NIST AI RMF 1.0\cite{ref1} & Partial & No & No & Guidance & No & Partial & Broad & High \\
    NCSC ML Security Principles\cite{ref2} & Partial & No & No & Principles & No & Partial & Broad & High \\
    CSA AI Model Risk Management\cite{ref4} & Partial & No & No & Gov. artefacts & No & Partial & Governance & High \\
    Microsoft AI Security Risk Assessment & Partial & Partial & Partial & Risk matrix & Partial & Partial & Broad & Med--High \\
    MITRE ATLAS\cite{ref3} & No & No & No & KB & Yes & No & Technique & Medium \\
    Bitton et al\cite{ref6}. & Partial & Partial & Partial & Scenario score & Partial & Partial & Lifecycle & High \\
    Grosse and Alahi\cite{ref7} & Partial & No & Partial & Severity score & Partial & Partial & Domain & Medium \\
    Yajima et al\cite{ref8} & Partial & No & Partial & Likelihood & No & Partial & Limited & Medium \\
    Our framework & Yes & Yes & Yes & F/I levels & Yes & Yes & System & Low \\
  \bottomrule
\end{tabular}%
}
\end{table}

\section{Methodology}

\subsection{Framework Overview and Problem Scope}

We present an evidence driven AI security risk assessment framework that transforms heterogeneous engineering artefacts into deterministic and auditable security decisions. The framework is implemented as a formally specified rule based decision system with explicit evaluation semantics. Each assessment is computed from bounded ordinal control strength scores and produces technique indexed feasibility and impact levels together with a trace that can be independently inspected and reproduced.

The workflow follows a Raw Data to Control IDs to Rules to Results pipeline. Let $R$ denote the set of externally inspectable artefacts produced during the development and operation of an AI system. Examples include repository configuration files, dependency manifests and lockfiles, CI and build logs, vulnerability and policy scan reports, provenance attestations, model and dataset documentation, and deployment configurations. All inputs to the framework are drawn from $R$ and are therefore evidence bound.

To enable comparability across projects, heterogeneous artefacts are normalised through a stable abstraction layer called Control IDs. A Control ID $c \in C$ denotes a security relevant property expressed in a project independent vocabulary. Each control definition specifies the artefact location $r_c \in R$ to be inspected and the rubric by which the observed state is translated into a discrete ordinal control strength score.

Control scoring is rubric guided and repeatable. For each $c \in C$, an assessor inspects the designated artefact and assigns a score
\[
s_c \in \{0,1,2,3\},
\]
where larger values indicate stronger control implementation. These values represent bounded control strength rather than probabilities, frequencies, or statistical estimates. Given a fixed rubric version and a fixed system snapshot, the resulting score vector
\[
\mathbf{s} = (s_c)_{c \in C}
\]
is fully determined by the observable artefacts in $R$.

Feasibility evaluation is performed by compiling technique level rule predicates that link control strength to adversarial capability. Let $T$ denote the set of techniques and $M$ the set of mitigations derived from a pinned MITRE ATLAS snapshot. The technique mitigation relation is represented as
\[
\mathcal{R} \subseteq T \times M.
\]

Because ATLAS mitigations are high level abstractions, a human specified mapping connects mitigations to the Control ID taxonomy. Each mitigation $m \in M$ is evaluated as a bounded mitigation strength
\[
\mu_m(\mathbf{s}) \in \{0,1,2,3\},
\]
computed deterministically from the underlying control strength scores.

For each technique $t \in T$, the compiler generates an ordered family of rule predicates over $\{\mu_m(\mathbf{s})\}$ and selects a discrete feasibility level
\[
F_t(\mathbf{s}) \in \{0,1,2,3\}
\]
under explicit descending priority semantics. Larger feasibility values indicate that the technique is more operationally achievable under the observed control strength configuration. Feasibility reflects adversarial capability conditional on the implemented controls and is not a probability, likelihood, or forecast of incident frequency.

Impact is modelled independently through the framework’s consequence parametrisation, yielding a technique indexed output pair
\[
(F_t(\mathbf{s}), I_t).
\]
Each reported outcome is accompanied by provenance metadata linking triggered predicates to contributing controls and ultimately to artefacts in $R$, ensuring full traceability from evidence to decision.

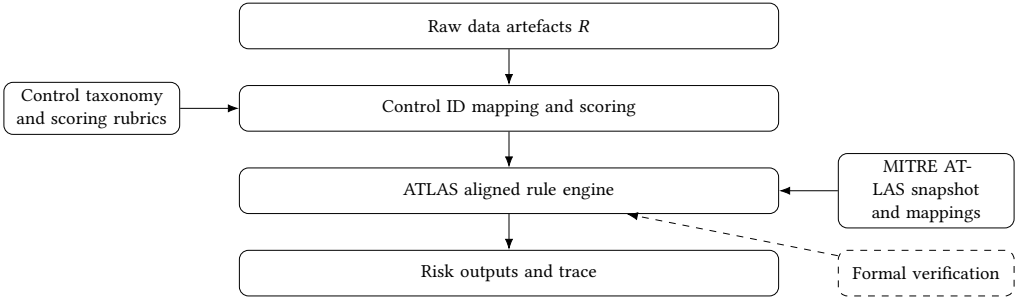
\begin{figure*}[!t]
\centering
\footnotesize
\resizebox{0.96\textwidth}{!}{%
\begin{tikzpicture}[
  >=Latex,
  node distance=5.5mm and 7mm
]
\tikzset{
  main/.style={
    rectangle, draw, rounded corners,
    align=center, minimum height=7mm,
    text width=0.58\textwidth
  },
  side/.style={
    rectangle, draw, rounded corners,
    align=center, minimum height=7mm,
    text width=0.18\textwidth
  },
  dashedbox/.style={
    rectangle, draw, dashed, rounded corners,
    align=center, minimum height=7mm,
    text width=0.18\textwidth
  }
}

\node[main] (raw) {Raw data artefacts $R$\\};
\node[main, below=of raw] (ctrl) {Control ID mapping and scoring};
\node[main, below=of ctrl] (engine) {ATLAS aligned rule engine\\};
\node[main, below=of engine] (out) {Risk outputs and trace};

\node[side, left=of ctrl, xshift=-2mm] (rubric) {Control taxonomy\\and scoring rubrics};
\node[side, right=of engine, xshift=2mm] (atlas) {MITRE ATLAS snapshot\\and mappings};
\node[dashedbox, right=of out, xshift=2mm] (fv) {Formal verification\\};

\draw[->] (raw) -- (ctrl);
\draw[->] (ctrl) -- (engine);
\draw[->] (engine) -- (out);

\draw[->] (rubric) -- (ctrl);
\draw[->] (atlas) -- (engine);
\draw[dashed,->] (fv) -- (engine);

\end{tikzpicture}%
}
\caption{End to End Workflow of the Evidence Driven Assessment Framework}
\label{fig:framework_workflow}
\end{figure*}

The scope of the framework is evidence based security assessment for AI enabled pipelines, with particular attention to lifecycle and supply chain exposures. It supports both vision and language model systems and remains extensible through additional Control IDs and updated ATLAS aligned mappings. Figure~\ref{fig:framework_workflow} illustrates the end to end workflow.

\subsection{Threat Model}

The framework is threat informed yet evidence bound. It does not estimate adversary intent or incident frequency. Instead, it evaluates whether a given adversarial technique is operationally feasible under the observable control strength configuration captured by artefact evidence. Feasibility therefore reflects adversarial capability conditional on implemented controls and must not be interpreted as probability, likelihood, or forecast of occurrence.

We model adversaries in terms of structural positions within the system boundary rather than specific identities. The assessment considers three representative attacker classes aligned with dominant system level exposures reflected in MITRE ATLAS techniques:

(i) a \emph{supply chain adversary} who can influence upstream dependencies, build inputs, model artefact distribution, or provenance metadata;

(ii) an \emph{API level attacker} who can interact with exposed model or service endpoints through standard request channels;

(iii) an \emph{insider with authorised data access} who can read, inject, or manipulate training or operational data within the permitted data pipeline.

These classes represent access positions. A given deployment may expose any subset of them. The API level attacker is restricted to black box interaction with deployed interfaces and cannot directly modify source code or configuration without an additional compromise. The supply chain adversary is constrained by development pipeline controls evidenced in artefacts, including version pinning, signing, provenance attestation, and vulnerability scanning. The insider is constrained by documented data governance mechanisms, access control enforcement, audit logging, and separation of duties where such controls are evidenced.

For each technique $t \in T$, the framework assumes that once the relevant access position is available, the adversary may repeatedly attempt the technique. The assessment therefore focuses on whether the control strength configuration $\mathbf{s}$ permits operational realisation of the technique, as captured by the feasibility function $F_t(\mathbf{s})$. It does not model attempt frequency, attacker utility, or strategic adaptation.

Within a single assessment run, the following elements are fixed: the system snapshot and artefact set $R$, the Control ID taxonomy and scoring rubrics, and the ATLAS derived compilation configuration, including the mitigation to control mapping $\Phi$ and technique profiles. Feasibility results are therefore conditional on a fixed policy configuration and a fixed evidence state.

\subsection{Raw Data Collection, Control ID Taxonomy, and Discrete Scoring}

The framework is grounded in raw data, denoted by a set of artefacts $R$, consisting of objective records produced during routine engineering and operational processes. Raw data refers to externally inspectable items such as configuration files, structured reports, and system logs rather than narrative statements of security posture. In our setting, $R$ is collected from four broad classes. The first class comprises software and system artefacts, including repository configuration, dependency manifests and lockfiles, container specifications, and release metadata. The second class comprises pipeline and assurance artefacts, including CI and build logs, vulnerability and policy scanning outputs, and automated compliance reports. The third class comprises provenance and supply chain artefacts, including bills of materials, attestations, signatures, and supplier governance records such as audit identifiers and review timestamps. The fourth class comprises deployment and operational artefacts, including API gateway configuration, authentication and authorisation settings, rate limiting policies, monitoring configuration, and rollback procedures. Across all classes, the defining property is that each item exists as a verifiable artefact that can be recollected and reinspected under a fixed system snapshot.

Direct use of $R$ for risk reasoning is impractical because raw artefacts are heterogeneous, tool dependent, and not directly comparable across projects. The framework therefore introduces Control IDs as an intermediate abstraction. A Control ID $c \in C$ denotes a security relevant property expressed in a project independent vocabulary. This abstraction serves two purposes. First, it enables maintainable rule logic by allowing rules to be written over a stable set of controls rather than file specific patterns. Second, it enables comparability and aggregation by reducing heterogeneous artefacts to a common set of security properties prior to feasibility computation. In this work, Control IDs are organised into major domains reflecting the assessment surface, including API and service controls, software and infrastructure supply chain controls, model and data provenance controls, and organisational governance controls. Collectively, the taxonomy captures exposure boundaries, abuse resistance, dependency integrity, reproducibility, provenance strength, and supplier governance.

Each Control ID is defined a priori by specifying the raw data location to be inspected and the rubric used to evaluate the observed state. Formally, each control binds to a designated artefact location $r_c \in R$. Control scoring is rubric guided and repeatable: the assessor locates $r_c$, inspects its objective state, and assigns a discrete control strength value according to the predefined rubric. Scores are assigned on a bounded four level ordinal scale
\[
s_c \in \{0,1,2,3\},
\]
where larger values indicate stronger implementation of the corresponding control. The scale is ordinal rather than numeric in nature; each level corresponds to an explicitly defined enforcement condition that can be justified by artefact evidence. The use of a discrete bounded scale avoids unwarranted numerical precision while preserving interpretability and auditability.

The control taxonomy, control to artefact bindings, and scoring rubrics are human specified at design time and therefore introduce a normative policy element. In addition, artefact inspection and rubric application are performed by a human assessor. Nevertheless, operational objectivity is preserved because outcomes are evidence bound and reproducible conditional on a fixed specification. Given the same artefact set $R$, the same control definitions, and the same rubric version, the resulting control strength vector $\mathbf{s}$ is determined by the observable content of the artefacts rather than discretionary interpretation. This separation confines subjectivity to an explicit and reviewable policy configuration while ensuring that assessment results can be re derived and audited against the underlying raw data.

The complete four level scoring rubric for the full Control ID taxonomy, with $|C| = 62$, including per control anchor examples, is provided in Appendix~\ref{app:rubric}.

\subsection{Rule Generation and Feasibility Computation}

Building on the evidence bound control abstraction, this stage transforms discrete control strength scores into an executable rule system aligned with established adversarial AI knowledge while remaining auditable against underlying artefacts. Purely handwritten technique rules are feasible for small rule bases but do not scale. Coverage becomes difficult to justify because the selection of techniques and mitigations is implicit, and maintenance becomes brittle when updates require manual modification of interdependent logic. To address these limitations, rule generation is grounded in MITRE ATLAS, a continuously updated knowledge base cataloguing adversary tactics, techniques, and mitigations for AI enabled systems. ATLAS provides a structured technique mitigation relation that supports systematic coverage and tractable updates via recompilation against a new snapshot.

Let $T$ denote the set of techniques and $M$ the set of mitigations. From ATLAS we extract a bipartite relation
\[
\mathcal{R} \subseteq T \times M
\]
where $(t,m) \in \mathcal{R}$ indicates that mitigation $m$ is associated with technique $t$. The relation $\mathcal{R}$ defines admissible structural dependencies between techniques and mitigations. Implementation strength is determined locally through control strength scores.

Because ATLAS mitigations are technology agnostic abstractions, mapping them to inspectable evidence requires an explicit normative specification. This specification is captured as part of a versioned policy object. We define a mitigation to control mapping function
\[
\Phi(m) = (A_m, B_m)
\]
where $A_m \subseteq C$ denotes mandatory controls and $B_m \subseteq C$ denotes compensating controls. Mandatory controls represent non substitutable requirements and are evaluated conjunctively. Compensating controls represent alternative mechanisms and are evaluated disjunctively. Once $\Phi$ and the control scoring rubrics are fixed, all subsequent computations are deterministic functions of the observable artefacts through the control strength vector.

Let $\mathbf{s} = (s_c)_{c \in C}$ denote the control strength vector. To preserve the bounded four level structure, we define a saturated control strength
\[
\bar{s}_c = \min(s_c, 3).
\]
Each mitigation $m \in M$ is compiled into a bounded mitigation strength
\[
\mu_m(\mathbf{s}) \in \{0,1,2,3\}
\]
defined by
\[
\mu_m(\mathbf{s}) = \min\big(g(A_m,\bar{\mathbf{s}}),\, h(B_m,\bar{\mathbf{s}})\big),
\]
where
\[
g(A_m,\bar{\mathbf{s}}) =
\begin{cases}
3, & A_m = \varnothing,\\
\min_{c \in A_m} \bar{s}_c, & \text{otherwise},
\end{cases}
\]
and
\[
h(B_m,\bar{\mathbf{s}}) =
\begin{cases}
3, & B_m = \varnothing,\\
\max_{c \in B_m} \bar{s}_c, & \text{otherwise}.
\end{cases}
\]
Mandatory requirements are therefore limited by the weakest element, while compensating evidence is credited according to the strongest available alternative.

For each technique $t \in T$, mitigations are partitioned into core and optional sets:
\[
K_t \subseteq \{ m \in M \mid (t,m) \in \mathcal{R} \},
\qquad
O_t \subseteq \{ m \in M \mid (t,m) \in \mathcal{R} \}.
\]
Core mitigations represent direct defences against the dominant attack mechanism and are treated as non compensable. Optional mitigations provide defence in depth or governance reinforcement. This partition is part of the policy configuration and is versioned explicitly.

Given $\{\mu_m(\mathbf{s})\}$, the compiler generates four ordered predicates per technique and evaluates a discrete feasibility level
\[
F_t(\mathbf{s}) \in \{0,1,2,3\},
\]
where larger values indicate greater operational feasibility under the observed control strength configuration. A representative template is:

\[
\begin{aligned}
\mathcal{P}_{t,3}(\mathbf{s}) &\equiv \exists\, m \in K_t,\ \mu_m(\mathbf{s}) = 0, \\
\mathcal{P}_{t,2}(\mathbf{s}) &\equiv
(\forall\, m \in K_t,\ \mu_m(\mathbf{s}) \ge 1)\ \wedge \\
&\quad
(\exists\, m \in K_t,\ \mu_m(\mathbf{s}) = 1\ \vee\ \exists\, m \in O_t,\ \mu_m(\mathbf{s}) = 0), \\
\mathcal{P}_{t,1}(\mathbf{s}) &\equiv
(\forall\, m \in K_t,\ \mu_m(\mathbf{s}) \ge 2)\ \wedge \\
&\quad
(\exists\, m \in K_t,\ \mu_m(\mathbf{s}) = 2\ \vee\ \exists\, m \in O_t,\ \mu_m(\mathbf{s}) = 1), \\
\mathcal{P}_{t,0}(\mathbf{s}) &\equiv
(\forall\, m \in K_t,\ \mu_m(\mathbf{s}) = 3)\ \wedge \\
&\quad
(\forall\, m \in O_t,\ \mu_m(\mathbf{s}) \ge 2).
\end{aligned}
\]

Feasibility is selected by descending priority evaluation:
\[
F_t(\mathbf{s}) =
\begin{cases}
3, & \mathcal{P}_{t,3}(\mathbf{s}),\\
2, & \mathcal{P}_{t,2}(\mathbf{s}),\\
1, & \mathcal{P}_{t,1}(\mathbf{s}),\\
0, & \mathcal{P}_{t,0}(\mathbf{s}).
\end{cases}
\]

This construction makes the semantics explicit. If any core mitigation is absent, feasibility is maximised. Intermediate feasibility levels reflect partial weakness in core mitigations or severe deficiencies in optional mitigations. The lowest feasibility level requires uniformly strong core mitigations and adequate optional reinforcement.

Impact is computed independently, and each technique yields an output pair $(F_t(\mathbf{s}), I_t)$. Separating feasibility from impact preserves interpretability because feasibility is entirely determined by evidence bound mitigation strength, while impact reflects stakeholder consequence modelling under a fixed parametrisation.

The overall method is a deterministic compilation pipeline. Human specified policy artefacts define $\Phi$ and $\{(K_t,O_t)\}_{t \in T}$, while program generated artefacts include mitigation strength functions and technique level predicates. Each evaluated predicate is accompanied by provenance metadata linking mitigations to contributing controls and ultimately to artefacts $r_c \in R$, ensuring full traceability from raw data to feasibility output.

\subsection{Impact Definition and Assignment}

We model impact as an ordinal consequence level that captures severity \emph{conditional on a technique succeeding} under a declared deployment context. Unlike feasibility, which is derived from repository observable control evidence through compiled predicates, impact is treated as a stakeholder anchored input that reflects consequence tolerance and recovery expectations. For each project $p$ and technique $t$, impact is represented as
\[
I_t^{(p)} \in \{0,1,2,3\},
\]
using the rubric in Table~\ref{tab:impact_rubric}.

In operational use, the deployment context $d^{(p)}$ and the corresponding impact profile $\{I_t^{(p)}\}_{t \in T}$ are provided or validated by the system owner and recorded within the versioned policy object $\mathcal{P}$. In this paper, we intentionally avoid context specific impact elicitation because real deployment contexts are not observable from repository artefacts for public open source projects. To preserve reproducibility and isolate evidence driven feasibility effects in our experiments, we fix a constant impact level for all techniques and projects:
\[
I_t^{(p)} = 2,\quad \forall t \in T,\ \forall p.
\]
Impacts are held fixed between baseline and hardened variants, so all reported changes arise solely from feasibility shifts under strengthened observable controls. Under this design, aggregate summary indices are interpretable as feasibility driven comparisons under a constant consequence assumption rather than as context calibrated consequence analysis.

When impact is uncertain or stakeholders disagree in practical deployments, we recommend recording a bounded range $(I_{t,\min}, I_{t,\max})$ and reporting conservatively using $I_{t,\max}$ for governance facing summaries, while optionally re running the analysis under $I_{t,\min}$ as a sensitivity check. This preserves auditability while avoiding false precision from contested impact assumptions.

\begin{table}[!t]
\caption{Ordinal impact rubric (0 lowest, 3 highest) used for technique consequence assignment.}
\label{tab:impact_rubric}
\centering
\footnotesize
\setlength{\tabcolsep}{5pt}
\renewcommand{\arraystretch}{1.15}
\resizebox{\textwidth}{!}{%
\begin{tabular}{cl}
\hline
Level & Anchor definition (consequence severity) \\
\hline
0 & Negligible: minimal harm with local and readily reversible effects. \\
1 & Limited: bounded disruption or minor degradation with limited external impact. \\
2 & Major: substantial service degradation, material integrity or confidentiality harm, or costly recovery. \\
3 & Severe: high consequence harm such as prolonged outage, large scale integrity failure, or systemic loss of trust. \\
\hline
\end{tabular}%
}
\end{table}

\subsection{Assessment Policy Configuration for Reproducibility and Auditability}

The framework includes normative configuration elements that cannot be inferred from artefacts alone. In particular, the mitigation to control mapping $\Phi$ and the technique profiles $\{(K_t,O_t)\}_{t \in T}$ determine how ATLAS mitigations are evidenced by the local Control ID taxonomy and how mitigation coverage is interpreted under the ordered predicate semantics. These elements define a governance policy rather than an empirical measurement. Once they are fixed, compilation and evaluation are fully deterministic functions of the observable artefacts through the control strength vector $\mathbf{s}$.

We therefore package all normative choices into a versioned assessment policy object
\[
\mathcal{P}=\langle \text{ATLAS snapshot},\ \mathcal{R},\ C \text{ with evidence bindings and rubrics},\ \Phi,\ \{(K_t,O_t)\}_{t \in T},\ \text{impact}\rangle,
\]
where $\mathcal{R}\subseteq T\times M$ is extracted from the selected ATLAS release, $\Phi(m)=(A_m,B_m)$ maps each mitigation to mandatory and compensating controls, and each technique profile partitions applicable mitigations into core $K_t$ and optional $O_t$ sets. Reproducing an assessment result requires a fixed project snapshot and a fixed policy. Concretely, the assessment report must uniquely identify the project snapshot, the Control ID taxonomy and its evidence bindings, the rubric version used to assign $s_c \in \{0,1,2,3\}$, and the exact policy object $\mathcal{P}$ that defines $\Phi$, $\{(K_t,O_t)\}$, and the impact configuration. In our experiments, the impact configuration fixes a constant impact level for all techniques and projects, $I_t^{(p)} = 2$, so that reported changes arise solely from feasibility differences.

The core and optional partition encodes mechanism criticality. A mitigation is classified as \emph{core} for technique $t$ when it blocks a necessary precondition or directly disrupts the dominant attack mechanism and is treated as non substitutable by design. A mitigation is classified as \emph{optional} when it provides defence in depth, governance support, or monitoring reinforcement that may reduce feasibility in intermediate bands but does not substitute for missing core protections. For example, for API mediated abuse and extraction techniques, access control and throttling are natural core candidates, while monitoring oriented controls are treated as optional. For supply chain compromise techniques, integrity anchoring controls such as version pinning, SBOM generation, and release hashing are natural core candidates, while response readiness controls are treated as optional.

The policy object $\mathcal{P}$ is maintained as a signed and versioned configuration bundle with an accompanying change log that records the rationale for modifications to $\Phi$ and $\{(K_t,O_t)\}_{t \in T}$. Any policy update triggers recompilation of the technique predicates and rerunning of the formal verification suite on the compiled rule set, ensuring that semantic guarantees hold for the released policy version. Each assessment report records a policy identifier and a content hash of the configuration bundle, enabling independent reproduction and audit of results against the same artefact snapshot and the same decision semantics.

\section{Formal Verification}
\subsection{Property selection}
The proposed framework operationalises security assessment as a deterministic decision function that maps structured evidence to discrete assessment outputs. Inputs take the form of bounded ordinal control strength scores, which are compiled into bounded mitigation strength values and evaluated by an ordered predicate system to yield a feasibility level for each technique. Such rule based evaluators are attractive because their decisions are transparent and auditable. However, they are also vulnerable to specification level faults that do not appear as runtime failures but as semantic failures. Examples include undefined outputs for admissible inputs, violations of intended priority semantics during rule selection, and counterintuitive behaviour in which stronger controls increase assessed feasibility. Because these faults concern the semantics of the decision function over a bounded but combinatorial input space, they cannot be established by empirical testing over a finite set of cases. We therefore introduce a formal verification layer to establish semantic correctness over the entire declared input domain.

We verify four properties that are necessary to treat the evaluator as a mathematically well defined and governance ready decision function. First, \emph{boundedness} ensures that all intermediate values, including mitigation strength and feasibility levels, remain within the declared four level scale. Without boundedness, out of range states would invalidate level interpretations and compromise any downstream aggregation or reporting. Second, \emph{totality} guarantees that every admissible control strength configuration yields a defined feasibility level for every technique. Without totality, the evaluator admits silent no decision cases that are unacceptable for operational assessment workflows. Third, \emph{ordered semantic consistency} formalises the intended ordered evaluation semantics. Under descending priority selection, correctness requires that the returned feasibility level is justified by its predicate and that no strictly higher level predicate is simultaneously satisfied. Without this property, results can contradict the declared precedence structure, undermining reproducibility. Fourth, \emph{monotonicity} captures the governance requirement induced by the control strength ordering: strengthening controls must not increase assessed feasibility. Violations create perverse incentives and render the evaluator internally inconsistent as a governance instrument. Together, these properties establish that the decision function is bounded, total, consistent under ordered rule evaluation, and aligned with the intended meaning of control strength.

\subsection{Verification procedure}

The verification objective is to show that, for all admissible control strength assignments $\mathbf{s}$ under a fixed policy configuration $\mathcal{P}$, the realised evaluator respects the declared discrete domain, is defined for every input, is consistent with the intended ordered semantics, and behaves monotonically under strengthened controls. Each property is checked by searching for counterexamples, namely assignments to $\mathbf{s}$ (or pairs $(\mathbf{s},\mathbf{s}')$) that violate the property. Because the input domain is explicitly bounded and discrete, these checks can be implemented as bounded model checking. Concretely, for each property we encode its negation as a satisfiability query over integer variables constrained to the finite range $\{0,1,2,3\}$ and invoke an SMT solver to search for a witness. If the query is satisfiable, the returned model provides a concrete counterexample that can be replayed against the executable predicates and rule evaluator. If the query is unsatisfiable, the property is discharged over the declared bounded input domain.

Boundedness is stated as a domain preservation condition for both intermediate mitigation strength values and the final feasibility outputs. Formally, it requires that for all admissible $\mathbf{s}$,
\[
\forall m \in M:\ \mu_m(\mathbf{s}) \in \{0,1,2,3\}
\quad \wedge \quad
\forall t \in T:\ F_t(\mathbf{s}) \in \{0,1,2,3\}.
\]
In counterexample form, the solver searches for an input $\mathbf{s}$ such that either $\exists m:\mu_m(\mathbf{s}) \notin \{0,1,2,3\}$ or $\exists t:F_t(\mathbf{s}) \notin \{0,1,2,3\}$. Discharging this property ensures that the evaluator never leaves the intended ordinal scale, which is a prerequisite for any downstream aggregation or summary index computation.

Totality states that the feasibility function is defined for every admissible input, meaning that the ordered predicate family provides at least one match for each technique. For a fixed technique $t$, totality can be expressed as
\begin{equation}
\forall \mathbf{s}:\ 
\mathcal{P}_{t,3}(\mathbf{s}) \,\vee\, \mathcal{P}_{t,2}(\mathbf{s}) \,\vee\, \mathcal{P}_{t,1}(\mathbf{s}) \,\vee\, \mathcal{P}_{t,0}(\mathbf{s}).
\end{equation}
Equivalently, the counterexample form searches for a ``no match'' assignment:
\begin{equation}
\exists \mathbf{s}:\ 
\neg \mathcal{P}_{t,3}(\mathbf{s}) \,\wedge\, 
\neg \mathcal{P}_{t,2}(\mathbf{s}) \,\wedge\, 
\neg \mathcal{P}_{t,1}(\mathbf{s}) \,\wedge\, 
\neg \mathcal{P}_{t,0}(\mathbf{s}).
\end{equation}
The absence of such an assignment establishes that the compiled rule set yields a feasibility output for every admissible input, eliminating silent no decision cases.

Ordered semantic consistency formalises that the implementation respects descending priority selection. In an ordered matching system, lower severity predicates may hold even when higher severity predicates hold. Correctness therefore requires that the selected level is witnessed by its predicate and that all strictly higher predicates are false at the same input. For each $k \in \{0,1,2,3\}$, this can be stated as
\begin{equation}
\forall \mathbf{s}:\ 
\bigl(F_t(\mathbf{s}) = k\bigr)\ \Rightarrow\
\Bigl(\mathcal{P}_{t,k}(\mathbf{s})\ \wedge\ \bigwedge_{j=k+1}^{3} \neg \mathcal{P}_{t,j}(\mathbf{s})\Bigr).
\end{equation}
The counterexample search targets any $\mathbf{s}$ where the evaluator returns $k$ but either $\mathcal{P}_{t,k}(\mathbf{s})$ is false or some $\mathcal{P}_{t,j}(\mathbf{s})$ with $j>k$ is true. Discharging this property rules out both miswitnessed outputs and priority inversions, establishing that the realised $F_t$ is a faithful implementation of the declared ordered semantics.

Monotonicity is verified as a relational property between two inputs, capturing the governance requirement induced by the control strength ordering. Let $\mathbf{s} \preceq \mathbf{s}'$ denote componentwise strengthening:
\begin{equation}
\mathbf{s} \preceq \mathbf{s}' \ \Longleftrightarrow\ \forall c \in C:\ s_c \le s'_c.
\end{equation}
The monotonicity requirement states that strengthening controls must not increase feasibility:
\begin{equation}
\forall \mathbf{s}, \mathbf{s}':\ (\mathbf{s} \preceq \mathbf{s}') \Rightarrow \bigl(F_t(\mathbf{s}') \le F_t(\mathbf{s})\bigr).
\end{equation}
Verification proceeds by searching for a counterexample pair $(\mathbf{s},\mathbf{s}')$ satisfying $\mathbf{s} \preceq \mathbf{s}'$ while violating the inequality, namely $F_t(\mathbf{s}') > F_t(\mathbf{s})$. The absence of a counterexample provides a formal guarantee that the evaluator does not exhibit perverse incentive behaviour under strengthened controls.

Across these properties, the procedure is uniform: each property is translated into a counterexample search problem over a finite domain. Any violation is witnessed by an explicit assignment, which supports direct debugging of predicates and iterative refinement of the compiled rule set.

\subsection{Verification results}

The verification run compiled the four semantic correctness properties into a finite set of proof obligations, each expressed as a counterexample search query over the bounded input domain. Table~\ref{tab:fv_summary} summarises the outcome. Across all properties, the solver returned no counterexample, discharging all obligations with a $100\%$ pass rate.

The $302$ proof obligations in Table~\ref{tab:fv_summary} correspond to solver queries generated at the level of compiled mitigation evaluators and technique level ordered decision semantics, rather than to individual concrete score assignments. Let $M$ denote the set of compiled mitigations and $T$ the set of techniques covered by the instantiated rule base after applying the policy configuration and Control ID taxonomy. Boundedness contributes one obligation per mitigation to show $\mu_m(\mathbf{s}) \in \{0,1,2,3\}$ and one per technique to show $F_t(\mathbf{s}) \in \{0,1,2,3\}$, yielding $|M| + |T|$ obligations. Totality contributes one obligation per technique that searches for a ``no match'' input where all four predicates $\mathcal{P}_{t,k}$ are false, yielding $|T|$ obligations. Ordered semantic consistency is checked per technique using two obligations, one that searches for miswitnessed outputs where $F_t(\mathbf{s}) = k$ but $\mathcal{P}_{t,k}(\mathbf{s})$ is false, and one that searches for priority inversions where $F_t(\mathbf{s}) = k$ while some $\mathcal{P}_{t,j}(\mathbf{s})$ with $j>k$ holds, yielding $2|T|$ obligations. Monotonicity contributes one relational obligation per technique that searches for a violating pair $(\mathbf{s},\mathbf{s}')$ with $\mathbf{s}\preceq\mathbf{s}'$ but $F_t(\mathbf{s}') > F_t(\mathbf{s})$, yielding $|T|$ obligations. In our instantiated rule base this gives
\[
(|M|+|T|) + |T| + 2|T| + |T| = |M| + 5|T| = 302,
\]
which matches the counts reported in Table~\ref{tab:fv_summary}.

These results provide semantic assurance for treating the evaluator as a well defined decision function over the declared bounded domain. Boundedness ensures that mitigation strength values and feasibility levels never leave the intended ordinal scale, preventing invalid states from propagating into any downstream aggregation or summary index computation. Totality guarantees that every admissible control strength configuration yields a defined feasibility level for every technique, ruling out silent no decision cases. Ordered semantic consistency confirms that the realised evaluator respects descending priority semantics, so each returned level is witnessed by its predicate and no strictly higher level predicate is simultaneously satisfied at the same input. Monotonicity ensures that strengthening controls cannot increase assessed feasibility, aligning decision behaviour with the intended interpretation of the control strength ordering.

Overall, the absence of counterexamples across $302$ obligations indicates that, for the current compiled rule base and bounded input domain, the framework satisfies a minimal set of semantic correctness requirements suitable for reproducible and auditable deployment.

\begin{table}[!t]
\caption{Formal verification summary over the bounded ordinal input domain.\label{tab:fv_summary}}
\centering
\footnotesize
\begin{tabular}{lrrrr}
\hline
Property & Obligations & Passed & Failed & Pass Rate \\
\hline
Boundedness & 86 & 86 & 0 & 100.00\% \\
Totality & 54 & 54 & 0 & 100.00\% \\
Ordered Semantic Consistency & 108 & 108 & 0 & 100.00\% \\
Monotonicity & 54 & 54 & 0 & 100.00\% \\
\hline
Overall & 302 & 302 & 0 & 100.00\% \\
\hline
\end{tabular}
\end{table}

\section{Experiment}
\subsection{Setup}

This experimental evaluation studies the framework as a deterministic, evidence driven decision function under two complementary intervention modes. The first is a controlled counterfactual unified hardening that sets a shared intervention set $C^{+}$ to a strong level in order to quantify the evaluator response under a standardised strengthening scenario. The second is a concrete evidence realised hardening in which selected projects are forked and specific security controls are genuinely implemented and committed as observable artefacts.

The study uses five public open source projects selected to span distinct AI enabled engineering profiles while keeping the evaluation compact and reproducible: KServe \cite{kserve}, vLLM \cite{vllm}, Ultralytics \cite{ultralytics}, TorchServe \cite{torchserve}, and SeldonIO cassava example \cite{cassava}. For each project $p$, we freeze the codebase to an explicitly identified Git snapshot so that evidence collection and scoring are reproducible. Table~\ref{tab:proj_snapshots} reports the snapshot identifier used for each repository, using a release tag where available and otherwise a commit hash, together with the snapshot date recorded at checkout time. All controls in the selected scope $C_{\mathrm{sel}}$ are scored against these frozen snapshots only.

We likewise freeze the external threat knowledge dependency by pinning a specific MITRE ATLAS snapshot used to extract the technique mitigation relation $\mathcal{R}$ and the derived mitigation set $M$. The ATLAS snapshot identifier and retrieval date are recorded in Table~\ref{tab:proj_snapshots} to avoid ambiguity under future ATLAS updates.

Technique coverage is determined mechanically by the instantiated policy object $\mathcal{P}$ and the evidence scope $C_{\mathrm{sel}}$ rather than by manual technique selection. Starting from the pinned ATLAS snapshot, we include a technique $t$ in the evaluated set $T$ if it has at least one listed mitigation $(t,m)\in\mathcal{R}$ for which the policy mapping $\Phi(m)$ references at least one control in the repository observable subset $C_{\mathrm{sel}}$. This yields $|T| = 54$ techniques in the present evaluation. Techniques with no mitigations mapped to $C_{\mathrm{sel}}$ under $\Phi$ are excluded because they cannot be assessed from snapshot bound repository artefacts without introducing unverifiable assumptions. For included techniques, if the designated artefact for a referenced control is absent or not observable in the frozen snapshot, the control is scored as missing with $s_c = 0$ under the rubric. This convention is conservative and increases feasibility rather than treating absent evidence as control presence. The evaluator remains total over the instantiated set $T$, so every included technique yields a defined feasibility level.

The full framework specification defines a Control ID taxonomy of $|C| = 62$ controls spanning multiple lifecycle phases. Not all corresponding raw data are expected to be present in a public Git snapshot, since many controls are evidenced through organisational and operational artefacts outside the codebase, such as internal design documentation, supplier due diligence packages, review records, runbooks, incident response exercises, or monitoring procedures. To preserve reproducibility and avoid unverifiable assumptions, we instantiate the assessment over a fixed subset of repository observable controls $C_{\mathrm{sel}}$ with $|C_{\mathrm{sel}}| = 15$. This selection constraint is experimental rather than conceptual. The complete four level scoring rubric for all Control IDs, including per control anchor examples, is provided in Appendix~\ref{app:rubric}.

The selected controls cover API and service governance, supply chain and dependency hygiene, vulnerability management practice, operational preparedness, and monitoring related controls. Each control $c \in C_{\mathrm{sel}}$ is scored on the ordinal scale $s_c \in \{0,1,2,3\}$ where larger values indicate stronger implementation. If the designated artefact for a control is not observable in the fixed snapshot, the control is treated as missing and assigned $s_c = 0$. This ensures that scoring is total and comparable across projects under a conservative evidence binding stance. To keep evidence binding concise and reproducible, Table~\ref{tab:ctrl_rawdata_fileset} specifies, for each Control ID, the admissible repository files used as evidence carriers. For a given project snapshot, the assessor checks the files in the corresponding set; if none are present or the required signal is not observable, the control is scored as missing.

To quantify actionability under a standardised improvement scenario, we define a counterfactual unified hardening condition applied uniformly across all projects over the same intervention set $C^{+}\subseteq C_{\mathrm{sel}}$. The hardening operator constructs a strengthened score vector $\mathbf{s'}^{(p)}$ by setting each control in $C^{+}$ to level $3$ while leaving all other controls unchanged:
\begin{equation}
s_c'^{(p)}=
\begin{cases}
3, & c\in C^{+},\\
s_c^{(p)}, & c\notin C^{+}.
\end{cases}
\end{equation}
Tables~\ref{tab:ctrl_scores_base} and~\ref{tab:ctrl_scores_harden} report the baseline control strength scores and the counterfactual changes under unified hardening.

Feasibility is then recomputed as $F_t(\mathbf{s'}^{(p)})$ while impacts are held fixed at $I_t^{(p)} = 2$ for all techniques and projects. The before and after comparison is therefore performed over $\{(F_t(\mathbf{s}^{(p)}), I_t^{(p)})\}$ and $\{(F_t(\mathbf{s'}^{(p)}), I_t^{(p)})\}$, isolating the evaluator response to strengthened observable controls under a constant consequence assumption.

To demonstrate that the framework responds to verifiable engineering changes rather than counterfactual score setting, we additionally fork two assessed projects and implement concrete controls in repository. The fork commits add an automated SBOM generation workflow together with the resulting SBOM artefact committed in repository, and a CI security analysis gate configured via a CodeQL workflow. For each fork, we recollect raw artefacts from the fork snapshot pinned to a specific commit, reassign Control ID scores using the admissible evidence carriers in Table~\ref{tab:ctrl_rawdata_fileset}, and rerun the full assessment pipeline. No manual score edits are applied, so all before and after score differences arise solely from newly observable fork artefacts.

To assess whether the rubric yields repeatable outcomes across assessors, we conduct a small interrater exercise. A second assessor independently rescored $C_{\mathrm{sel}}$ for two projects using only the admissible evidence carriers in Table~\ref{tab:ctrl_rawdata_fileset}$,$ at the pinned commits in Table~\ref{tab:proj_snapshots}. We report per control comparisons (Table~\ref{tab:interrater_scores}) and agreement statistics (Table~\ref{tab:interrater_agreement}). Disagreements are resolved by adjudication in which both assessors reinspect the same admissible evidence; when evidence remains ambiguous, the lower score is retained to preserve auditability.

\begin{table}[!t]
\caption{Pinned Snapshots Used in the Evaluation (Reproducibility Manifest)}
\label{tab:proj_snapshots}
\centering
\footnotesize
\setlength{\tabcolsep}{4pt}
\renewcommand{\arraystretch}{1.15}
\begin{tabular}{p{2.25cm}p{1.55cm}p{2cm}p{2.15cm}}
\hline
Item & Type & Identifier & Date \\
\hline
KServe & commit & \texttt{5b033a4} & \texttt{2025-11-3} \\
vLLM & commit & \texttt{b17039b} & \texttt{2026-01-20} \\
Ultralytics &commit & \texttt{c529572} & \texttt{2026-01-13} \\
TorchServe & tag & \texttt{v0.0.9} & \texttt{2024-07-30} \\
cassava-example & tag/commit & \texttt{-} & \texttt{-} \\
MITRE ATLAS & tag & \texttt{5.1.1} & \texttt{2025-11-26} \\
\hline
\end{tabular}
\end{table}

% ---------------------------
% Table 1: Baseline scores only (compact)
% ---------------------------
\begin{table*}[!t]
\caption{Baseline Control Scores (0 Weakest, 3 Strongest). '---' = not applicable.}
\label{tab:ctrl_scores_base}
\centering
\footnotesize
\setlength{\tabcolsep}{4pt}
\renewcommand{\arraystretch}{1.15}
\resizebox{\textwidth}{!}{%
\begin{tabular}{lccccc}
\hline
Control ID & KServe & vLLM & Ultralytics & TorchServe & cassava-example \\
\hline
\texttt{api.exposure}                 & 2 & 2 & ---  & 1 & 0 \\
\texttt{api.access\_control}          & 2 & 1 & ---  & 1 & 0 \\
\texttt{api.rate\_limit}              & 1 & 0 & ---  & 0 & 0 \\
\texttt{api.abuse\_detection}         & 0 & 0 & ---  & 0 & 0 \\
\texttt{sc.version\_pinning}          & 3 & 1 & 1 & 0 & 0 \\
\texttt{sc.sbom}                      & 0 & 0 & 0 & 0 & 0 \\
\texttt{model.hash\_release}          & 1 & 1 & 1 & 1 & 0 \\
\texttt{sec.vulnerability\_scan}      & 0 & 2 & 2 & 0 & 0 \\
\texttt{org.vdp}                      & 2 & 2 & 2 & 1 & 0 \\
\texttt{org.incident\_response}       & 1 & 1 & 2 & 0 & 0 \\
\texttt{org.rollback}                 & 2 & 0 & 1 & 1 & 0 \\
\texttt{org.threat\_modeling}         & 0 & 0 & 0 & 0 & 0 \\
\texttt{org.asset\_inventory}         & 0 & 0 & 1 & 0 & 0 \\
\texttt{org.audit\_trail}             & 0 & 0 & 1 & 0 & 0 \\
\texttt{eval.performance\_monitoring} & 2 & 2 & ---  & 2 & 0 \\
\hline
\end{tabular}%
}
\end{table*}

% ---------------------------
% Table 2: Only hardened controls (C+) show s -> s'
% ---------------------------
\begin{table*}[!t]
\caption{Unified Hardening Effects for the Intervention Set $C^{+}$}
\label{tab:ctrl_scores_harden}
\centering
\footnotesize
\setlength{\tabcolsep}{5pt}
\renewcommand{\arraystretch}{1.15}
\resizebox{\textwidth}{!}{%
\begin{tabular}{lccccc}
\hline
Control ID ($c\in C^{+}$) & KServe ($s\!\rightarrow\!s'$) & vLLM ($s\!\rightarrow\!s'$) & Ultralytics ($s\!\rightarrow\!s'$) & TorchServe ($s\!\rightarrow\!s'$) & cassava-example ($s\!\rightarrow\!s'$) \\
\hline
\texttt{sc.version\_pinning}          & 3$\rightarrow$3 & 1$\rightarrow$3 & 1$\rightarrow$3 & 0$\rightarrow$3 & 0$\rightarrow$3 \\
\texttt{sec.vulnerability\_scan}      & 0$\rightarrow$3 & 2$\rightarrow$3 & 2$\rightarrow$3 & 0$\rightarrow$3 & 0$\rightarrow$3 \\
\texttt{sc.sbom}                      & 0$\rightarrow$3 & 0$\rightarrow$3 & 0$\rightarrow$3 & 0$\rightarrow$3 & 0$\rightarrow$3 \\
\texttt{model.hash\_release}          & 1$\rightarrow$3 & 1$\rightarrow$3 & 1$\rightarrow$3 & 1$\rightarrow$3 & 0$\rightarrow$3 \\
\texttt{org.vdp}                      & 2$\rightarrow$3 & 2$\rightarrow$3 & 2$\rightarrow$3 & 1$\rightarrow$3 & 0$\rightarrow$3 \\
\texttt{org.incident\_response}       & 1$\rightarrow$3 & 1$\rightarrow$3 & 2$\rightarrow$3 & 0$\rightarrow$3 & 0$\rightarrow$3 \\
\texttt{org.rollback}                 & 2$\rightarrow$3 & 0$\rightarrow$3 & 1$\rightarrow$3 & 1$\rightarrow$3 & 0$\rightarrow$3 \\
\texttt{eval.performance\_monitoring} & 2$\rightarrow$3 & 2$\rightarrow$3 & ---             & 2$\rightarrow$3 & 0$\rightarrow$3 \\
\hline
\end{tabular}%
}
\end{table*}

\begin{table*}[!t]
\caption{Raw Data File Sets Used to Bind Each Control ID}
\label{tab:ctrl_rawdata_fileset}
\centering
\footnotesize
\setlength{\tabcolsep}{6pt}
\renewcommand{\arraystretch}{1.2}
\resizebox{\textwidth}{!}{%
\begin{tabular}{lp{0.86\textwidth}}
\hline
Control ID & Evidence file set (admissible raw data carriers) \\
\hline
\texttt{api.exposure} &
\texttt{README.md}, \texttt{QUICKSTART.md}, \texttt{docs}. \\
\texttt{api.access\_control} &
\texttt{README.md}, \texttt{SECURITY.md}, \texttt{docs}. \\
\texttt{api.rate\_limit} &
\texttt{README.md}, \texttt{docs}, \texttt{values.yaml}, \texttt{Chart.yaml}. \\
\texttt{api.abuse\_detection} &
\texttt{SECURITY.md}, \texttt{README.md}, \texttt{docs}. \\
\hline
\texttt{sc.version\_pinning} &
\texttt{requirements.txt}, \texttt{requirements.lock}, \texttt{poetry.lock}, \texttt{Pipfile.lock}, \texttt{pyproject.toml}, \texttt{package-lock.json}, \texttt{pnpm-lock.yaml}, \texttt{yarn.lock}, \texttt{go.mod}, \texttt{go.sum}. \\
\texttt{sc.sbom} &
\texttt{sbom.spdx}, \texttt{sbom.json}, \texttt{sbom.xml}, \texttt{bom.json}, \texttt{SPDX}. \\
\texttt{model.hash\_release} &
\texttt{SHA256SUMS}, \texttt{SHA256SUMS.txt}, \texttt{checksums.txt}, \texttt{*.sha256}, \texttt{*.sha512}, \texttt{*.sig}. \\
\texttt{sec.vulnerability\_scan} &
\texttt{codeql.yml}, \texttt{codeql-analysis.yml}, \texttt{dependabot.yml}, \texttt{snyk.yml}, \texttt{trivy.yml}, \texttt{grype.yml}, \texttt{SECURITY.md}. \\
\hline
\texttt{org.vdp} &
\texttt{SECURITY.md}, \texttt{.github/SECURITY.md}. \\
\texttt{org.incident\_response} &
\texttt{SECURITY.md}, \texttt{INCIDENT\_RESPONSE.md}, \texttt{IR.md}. \\
\texttt{org.rollback} &
\texttt{README.md}, \texttt{DEPLOYMENT.md}, \texttt{RUNBOOK.md}, \texttt{OPERATIONS.md}. \\
\texttt{org.threat\_modeling} &
\texttt{THREAT\_MODEL.md}, \texttt{threatmodel.md}, \texttt{SECURITY.md}. \\
\texttt{org.asset\_inventory} &
\texttt{ARCHITECTURE.md}, \texttt{COMPONENTS.md}, \texttt{docs}. \\
\texttt{org.audit\_trail} &
\texttt{AUDIT.md}, \texttt{LOGGING.md}, \texttt{README.md}. \\
\texttt{eval.performance\_monitoring} &
\texttt{METRICS.md}, \texttt{MONITORING.md}, \texttt{README.md}. \\
\hline
\end{tabular}%
}
\end{table*}

% ---------------------------
% Table X: Inter-rater scoring comparison (Assessor A vs Assessor B)
% ---------------------------
\begin{table*}[!t]
\caption{Interrater control scoring comparison on two frozen snapshots. A=Assessor A; B=Assessor B}
\label{tab:interrater_scores}
\centering
\footnotesize
\setlength{\tabcolsep}{4pt}
\renewcommand{\arraystretch}{1.12}
\begin{tabular}{lcccc}
\hline
Control ID & KServe (A) & KServe (B) & vLLM (A) & vLLM (B) \\
\hline
\texttt{api.exposure}                & 2 & 2 & 2 & 2 \\
\texttt{api.access\_control}         & 2 & 1 & 1 & 1 \\
\texttt{api.rate\_limit}             & 1 & 0 & 0 & 0 \\
\texttt{api.abuse\_detection}        & 0 & 1 & 0 & 0 \\
\texttt{sc.version\_pinning}         & 3 & 3 & 1 & 1 \\
\texttt{sc.sbom}                     & 0 & 0 & 0 & 0 \\
\texttt{model.hash\_release}         & 1 & 0 & 1 & 0 \\
\texttt{sec.vulnerability\_scan}     & 0 & 2 & 2 & 1 \\
\texttt{org.vdp}                     & 2 & 3 & 2 & 2 \\
\texttt{org.incident\_response}      & 1 & 2 & 1 & 2 \\
\texttt{org.rollback}                & 2 & 2 & 0 & 0 \\
\texttt{org.threat\_modeling}        & 0 & 0 & 0 & 1 \\
\texttt{org.asset\_inventory}        & 0 & 1 & 0 & 0 \\
\texttt{org.audit\_trail}            & 0 & 0 & 0 & 0 \\
\texttt{eval.performance\_monitoring}& 2 & 2 & 2 & 2 \\
\hline
\end{tabular}
\end{table*}

\subsection{Metrics}

Metrics quantify the magnitude and structural characteristics of assessment change induced by hardening, using only the computed feasibility and impact levels. For each project $p$ and technique $t$, we define a technique level feasibility impact contribution
\begin{equation}
U_t^{(p)} = F_t(\mathbf{s}^{(p)})\, I_t^{(p)}, \qquad
U_t'^{(p)} = F_t(\mathbf{s}'^{(p)})\, I_t^{(p)}.
\end{equation}
Because feasibility and impact are reported on a four level ordinal scale, any scalar aggregation is interpreted as a bounded heuristic summary rather than a calibrated real world risk measure.

We introduce a Feasibility Impact Summary Index (FISI) to compare project states under a fixed assessment policy:
\begin{equation}
\mathrm{FISI}(p)=\sum_{t\in T} U_t^{(p)}, \qquad
\mathrm{FISI}'(p)=\sum_{t\in T} U_t'^{(p)}.
\end{equation}
The effect of hardening is reported via absolute and relative reductions:
\begin{equation}
\Delta \mathrm{FISI}(p)=\mathrm{FISI}(p)-\mathrm{FISI}'(p),
\end{equation}
\begin{equation}
r\Delta \mathrm{FISI}(p)=\frac{\mathrm{FISI}(p)-\mathrm{FISI}'(p)}{\mathrm{FISI}(p)+\epsilon},
\end{equation}
where $\epsilon$ is a small constant used only to avoid division by zero.

To capture governance relevant worst case behaviour, we report the maximum technique contribution before and after hardening:
\begin{equation}
WCR(p)=\max_{t\in T} U_t^{(p)},
\qquad
WCR'(p)=\max_{t\in T} U_t'^{(p)}.
\end{equation}

We additionally measure the density of techniques in a high feasibility and high impact region under fixed thresholds $\tau_F$ and $\tau_I$. We set $\tau_F=\tau_I=2$ so that the region corresponds to levels $2$ and $3$ on the four level ordinal scale:
\begin{equation}
HRS(p)=\frac{\left|\left\{t\in T:\ F_t(\mathbf{s}^{(p)})\ge\tau_F\ \wedge\ I_t^{(p)}\ge\tau_I\right\}\right|}{|T|},
\end{equation}
\begin{equation}
HRS'(p)=\frac{\left|\left\{t\in T:\ F_t(\mathbf{s}'^{(p)})\ge\tau_F\ \wedge\ I_t^{(p)}\ge\tau_I\right\}\right|}{|T|}.
\end{equation}
In our experiments, impact is fixed as $I_t^{(p)}=2$ for all $t$ and $p$, so the impact threshold is always satisfied and $HRS$ reduces to the fraction of techniques with feasibility at least $2$.

To characterise structural change, we normalise technique contributions into a bounded profile:
\begin{equation}
q_t^{(p)}=\frac{U_t^{(p)}}{\sum_{j\in T} U_j^{(p)}+\epsilon},
\qquad
q_t'^{(p)}=\frac{U_t'^{(p)}}{\sum_{j\in T} U_j'^{(p)}+\epsilon},
\end{equation}
where the same $\epsilon$ convention avoids undefined normalisation when all contributions are zero. We then compute an entropy based concentration score:
\begin{equation}
H(p)=-\sum_{t\in T} q_t^{(p)}\log q_t^{(p)},
\qquad
H'(p)=-\sum_{t\in T} q_t'^{(p)}\log q_t'^{(p)},
\end{equation}
using the standard convention $0\log 0 = 0$. Lower entropy indicates that contributions are concentrated in fewer techniques, whereas higher entropy indicates a more diffuse profile.

Finally, we quantify divergence between the normalised profiles using Jensen Shannon divergence. Let $m=\tfrac{1}{2}\bigl(q^{(p)}+q'^{(p)}\bigr)$; then
\begin{equation}
JSD(p)=\tfrac{1}{2}KL(q^{(p)}\|m)+\tfrac{1}{2}KL(q'^{(p)}\|m),
\end{equation}
where $KL(\cdot\|\cdot)$ denotes Kullback Leibler divergence computed with the same $0\log 0=0$ convention. This bounded and symmetric measure captures how strongly each project’s contribution profile responds to the same hardening intervention, enabling direct comparison of sensitivity across case studies.

\section{Results}
\subsection{Overall Hardening Effect Across Projects}

Table~\ref{tab:overall_metrics} summarises outcomes before and after unified hardening over the intervention set $C^{+}$. We treat ordinal preserving metrics as primary evidence of change and use FISI as a secondary summary aid. Because impact is fixed at $I_t^{(p)}=2$ for all techniques and projects, $WCR$ and $HRS$ primarily reflect shifts in feasibility rather than changes in consequence assumptions.

First, the worst case contribution $WCR$ remains at $6$ for all projects. Under $U_t^{(p)} = F_t(\mathbf{s}^{(p)}) I_t^{(p)}$ with $I_t^{(p)}=2$, a value of $6$ implies that at least one technique still attains feasibility level $F_t=3$ after hardening. The unified hardening package therefore reduces aggregate contributions across many techniques but does not remove the most severe residual feasibility cases that dominate the maximum.

Second, the high region share $HRS$ decreases for four projects and remains unchanged for KServe. With $\tau_F=\tau_I=2$ and $I_t^{(p)}=2$ fixed, $HRS$ reduces to the fraction of techniques with feasibility at least $2$. Under this thresholded view, only improvements that cross the feasibility boundary from $2$ to $1$ or from $3$ to $1$ reduce $HRS$. For example, Ultralytics decreases from $0.9815$ to $0.8333$, TorchServe from $0.9259$ to $0.8333$, and Cassava Example from $1.0000$ to $0.8333$, whereas KServe remains at $0.5185$ despite a noticeable reduction in FISI. This pattern is consistent with feasibility improvements that occur within the same feasibility band and therefore do not change the threshold count.

As a secondary summary, FISI decreases for all projects, indicating broad downward shifts in feasibility when aggregated across techniques under the fixed policy. Entropy $H$ decreases modestly in all cases and the Jensen Shannon divergence values remain small, suggesting that unified hardening primarily produces a coherent reduction across the contribution profile while leaving a small set of dominant residual techniques that continue to shape worst case behaviour.

\begin{table*}[!t]
\caption{Aggregate Assessment Metrics Before and After Unified Hardening}
\label{tab:overall_metrics}
\centering
\footnotesize
\resizebox{\textwidth}{!}{%
\begin{tabular}{lrrrrrrrrrrr}
\hline
Project & FISI$_b$ & FISI$_a$ & $\Delta$FISI & $r\Delta$FISI & WCR$_b$ & WCR$_a$ & HRS$_b$ & HRS$_a$ & H$_b$ & H$_a$ & JSD \\
\hline
KServe & 198.00 & 184.00 & 14.00 & 0.0707 & 6.00 & 6.00 & 0.5185 & 0.5185 & 3.8366 & 3.7055 & 0.0252 \\
vLLM & 232.00 & 210.00 & 22.00 & 0.0948 & 6.00 & 6.00 & 0.6852 & 0.6481 & 3.8999 & 3.7372 & 0.0340 \\
Ultralytics & 298.00 & 256.00 & 42.00 & 0.1409 & 6.00 & 6.00 & 0.9815 & 0.8333 & 3.9726 & 3.7978 & 0.0515 \\
TorchServe & 268.00 & 230.00 & 38.00 & 0.1418 & 6.00 & 6.00 & 0.9259 & 0.8333 & 3.9531 & 3.7874 & 0.0444 \\
Cassava Example & 324.00 & 266.00 & 58.00 & 0.1790 & 6.00 & 6.00 & 1.0000 & 0.8333 & 3.9890 & 3.8039 & 0.0622 \\
\hline
\end{tabular}%
}
\end{table*}

\subsection{Interrater consistency}

Table~\ref{tab:interrater_scores} compares control strength scores assigned by two assessors for two frozen snapshots, yielding 30 control instances. Exact agreement is 60.0\%, while within one level agreement reaches 96.7\%, indicating that most disagreements are confined to adjacent ordinal levels rather than reflecting systematic reinterpretation of the rubric. Because the 0 to 3 scale is ordinal, we additionally report weighted agreement. Cohen's $\kappa$ is 0.58 under linear weights and 0.73 under quadratic weights (Table~\ref{tab:interrater_agreement}), which is consistent with moderate to substantial agreement.

Disagreements concentrate in cases where evidence exists as narrative documentation but lacks a dedicated, inspectable artefact in the pinned snapshot. Two recurring patterns are observed. First, pipeline controls such as \texttt{sec.vulnerability\_scan} and related security analysis practices can be described in documentation without being enforced as a CI gate, creating ambiguity between documented intent and automated enforcement. Second, release integrity controls such as \texttt{model.hash\_release} depend on whether checksum or signature artefacts are committed or published in an inspectable location, which can vary across releases and is not always represented in the repository snapshot. Similar ambiguity arises for governance oriented controls, for example whether a vulnerability disclosure channel is present as a formal policy artefact or only implied by informal guidance.

Following adjudication, ambiguous cases are conservatively assigned the lower score. This resolution rule aligns with the framework's evidence binding stance by avoiding the interpretation of absent or weakly evidenced signals as control presence. Importantly, the observed disagreement structure suggests that assessor variation is unlikely to dominate downstream feasibility outputs: since nearly all differences are within one ordinal level, they tend to perturb mitigation strength values locally rather than inducing large, systematic shifts across many techniques. We therefore treat the published rubric and admissible evidence carrier list as sufficiently stable for snapshot bound studies while noting that, for controls that distinguish intent from enforcement, additional artefact anchors would further improve assessor agreement.

% ---------------------------
% Table Y: Inter-rater agreement summary
% ---------------------------
\begin{table}[!t]
\caption{Interrater agreement summary over 30 control instances (15 controls $\times$ 2 projects). Weighted $\kappa$ accounts for ordinal distance on the 0 to 3 scale.}
\label{tab:interrater_agreement}
\centering
\footnotesize
\setlength{\tabcolsep}{6pt}
\renewcommand{\arraystretch}{1.15}
\begin{tabular}{lr}
\hline
Metric & Value \\
\hline
Exact agreement & 60.0\% \\
Within one level agreement & 96.7\% \\
Cohen's \(\kappa\) (unweighted) & 0.41 \\
Cohen's \(\kappa\) (linear-weighted) & 0.58 \\
Cohen's \(\kappa\) (quadratic-weighted) & 0.73 \\
\hline
\end{tabular}
\end{table}

\subsection{Sensitivity to Alternative Policy Mappings}

The assessment policy $\mathcal{P}$ includes normative mapping choices, particularly the mitigation to control composition $\Phi$ that determines how multiple controls jointly witness a mitigation. To test whether our empirical conclusions depend on a single mapping interpretation, we perform a small sensitivity study that fixes the same frozen snapshots, Control ID scores, and unified hardening set $C^{+}$, and varies only $\Phi$ across three reasonable governance stances.

We evaluate three mapping variants. Base uses the original mandatory and compensating structure, where $\Phi(m)=(A_m,B_m)$ and mitigation strength is computed as
\[
\mu_m(\mathbf{s})=\min\bigl(\min_{c\in A_m}\bar{s}_c,\ \max_{c\in B_m}\bar{s}_c\bigr),
\]
with the empty set conventions defined in Section 3.
Strict treats all mapped controls as non substitutable by collapsing the witness set to conjunction, yielding
\[
\mu_m^{\mathrm{strict}}(\mathbf{s})=\min_{c\in A_m\cup B_m}\bar{s}_c.
\]
Permissive treats all mapped controls as substitutable by collapsing the witness set to disjunction, yielding
\[
\mu_m^{\mathrm{perm}}(\mathbf{s})=\max_{c\in A_m\cup B_m}\bar{s}_c.
\]
All other elements of $\mathcal{P}$ are held fixed, including the technique profiles $\{(K_t,O_t)\}$, the technique set $T$, and the constant impact assignment $I_t^{(p)}=2$.

Table~\ref{tab:policy_sensitivity} shows that the headline conclusion is robust under Base and Permissive mappings. Unified hardening reduces FISI for all five projects, with $r\Delta \mathrm{FISI}$ spanning $0.0707$ to $0.1790$ under Base and increasing to $0.1042$ to $0.2346$ under the more substitutable Permissive policy. Larger reductions under Permissive are expected because substitutability increases the chance that strengthening any single control raises the realised mitigation strength, which can reduce feasibility across multiple techniques.

In contrast, Strict collapses to a saturated regime in which hardening produces no change. Under Strict, Table~\ref{tab:policy_sensitivity} reports $\mathrm{FISI}_b=\mathrm{FISI}_a=324$ for all projects and $r\Delta \mathrm{FISI}=0$. Since $|T|=54$ and $I_t^{(p)}=2$, the value $324$ equals $|T|\cdot 3 \cdot 2$, implying that feasibility is maximised at $F_t=3$ for all techniques. Under an all controls minimum composition, any unmoved weak control acts as a bottleneck, preventing improvements in strengthened controls from propagating to mitigation strength and downstream feasibility. This illustrates a practical loss of discriminative power under an overly conservative non substitutability assumption for the chosen evidence scope.

Policy strictness is a governance decision. More conservative mappings may be preferred in high assurance contexts, while more permissive mappings may be acceptable when evidence scope is narrow or when compensating mechanisms are operationally valid. However, calibration is necessary to preserve analytical usefulness and to avoid degenerate configurations that saturate feasibility and become insensitive to observable improvements.

We recommend a simple saturation check on a small calibration set of representative projects under baseline conditions. Let
\[
S_3(p)=\frac{\left|\left\{t\in T:\ F_t(\mathbf{s}^{(p)})=3\right\}\right|}{|T|}
\]
denote the fraction of techniques assigned to the maximum feasibility level. If $S_3(p)$ exceeds a high threshold for most calibration projects, for example $S_3(p)>X$ with $X=0.9$, then the policy is too strict for the current evidence scope because it forces most techniques into the maximum band regardless of project state. Symmetrically, one may monitor
\[
S_0(p)=\frac{\left|\left\{t\in T:\ F_t(\mathbf{s}^{(p)})=0\right\}\right|}{|T|}
\]
to detect overly permissive settings that collapse most techniques to the minimum band. Overall, this sensitivity study shows that the observed hardening effect is not an artefact of a single mapping choice, while highlighting that practical policy calibration must balance conservatism with the ability to distinguish and reward meaningful control improvements.

% ---------------------------
% Sensitivity / Ablation Table: Policy mapping variants
% ---------------------------
\begin{table*}[!t]
\caption{Sensitivity of hardening outcomes to alternative policy mapping variants.}
\label{tab:policy_sensitivity}
\centering
\footnotesize
\setlength{\tabcolsep}{5pt}
\renewcommand{\arraystretch}{1.15}
\resizebox{\textwidth}{!}{%
\begin{tabular}{lrrr|rrr|rrr}
\hline
& \multicolumn{3}{c|}{Base $\Phi$} & \multicolumn{3}{c|}{Strict $\Phi$} & \multicolumn{3}{c}{Permissive $\Phi$} \\
Project & $FISI_b$ & $FISI_a$ & $r\Delta FISI$ & $FISI_b$ & $FISI_a$ & $r\Delta FISI$ & $FISI_b$ & $FISI_a$ & $r\Delta FISI$ \\
\hline
KServe            & 198.00 & 184.00 & 0.0707 & 324.00 & 324.00 & 0.0000 & 192.00 & 172.00 & 0.1042 \\
vLLM              & 232.00 & 210.00 & 0.0948 & 324.00 & 324.00 & 0.0000 & 230.00 & 200.00 & 0.1304 \\
Ultralytics       & 298.00 & 256.00 & 0.1409 & 324.00 & 324.00 & 0.0000 & 284.00 & 238.00 & 0.1620 \\
TorchServe        & 268.00 & 230.00 & 0.1418 & 324.00 & 324.00 & 0.0000 & 260.00 & 216.00 & 0.1692 \\
cassava example   & 324.00 & 266.00 & 0.1790 & 324.00 & 324.00 & 0.0000 & 324.00 & 248.00 & 0.2346 \\
\hline
\end{tabular}%
}
\end{table*}

\subsection{Fork based Concrete Control Implementation}

Unified hardening quantifies how feasibility would change under a standardised strengthening scenario applied in the abstract. We next validate that the framework responds to real and auditable engineering changes by evaluating forked project variants in which specific controls are genuinely implemented and committed as observable artefacts. Concretely, we implement two controls in repository: automated SBOM generation (Control ID \texttt{sc.sbom}) and a CI security analysis gate via CodeQL (Control ID \texttt{sec.vulnerability\_scan}). Table~\ref{tab:fork_manifest} pins the fork commits used for re assessment, and Table~\ref{tab:fork_ctrl_delta} reports the induced Control ID score changes derived purely from observable evidence. No manual score edits are applied.

The implemented changes are intentionally minimal and targeted. SBOM generation improves evidence of dependency inventory and supply chain visibility, while CodeQL gating strengthens automated vulnerability analysis enforcement in CI. Under our policy mapping, these controls contribute to mitigation strength values associated with supply chain compromise and code level weakness exploitation techniques, which in turn can reduce feasibility for techniques whose core or optional mitigations are witnessed by these controls. This provides a direct end to end test that links committed artefacts to control strength scores, mitigation strength values, and technique level feasibility outcomes.

Table~\ref{tab:fork_feasibility} summarises technique feasibility distributions before and after the fork implementations. For both KServe and vLLM, the distribution shift is small but directionally consistent with improved control evidence. In KServe, one technique moves from feasibility level $F=1$ to $F=0$, increasing the count at $F=0$ from 2 to 3 while leaving the $F=3$ and $F=2$ bands unchanged. In vLLM, one technique similarly shifts from $F=1$ to $F=0$. These changes are expected given that only two Control IDs are strengthened and that many techniques in the evaluated set are driven by other controls that remain unchanged. Importantly, the shifts occur in the lowest band boundary, indicating that the added artefacts are sufficient to satisfy at least one additional best case predicate under the ordered semantics for a small number of techniques.

The absence of change in the $F=3$ band is also informative. Since $F=3$ is triggered by missing core mitigations, a minimal intervention that improves only SBOM and CI scanning is unlikely to eliminate techniques whose feasibility is dominated by other missing core controls, such as access control, rate limiting, abuse detection, or organisational governance artefacts that are outside the repository evidence scope. This mirrors the earlier unified hardening result in which worst case behaviour remains unchanged, and it highlights that reducing maximum feasibility requires addressing technique specific core controls rather than only improving general hygiene controls.

Comparing fork based changes with unified hardening clarifies the role of evidence realisation. Unified hardening sets multiple controls to level 3 simultaneously and therefore provides an upper bound response under the same policy. The fork experiment, by contrast, implements a narrow subset of controls with realistic strength levels and produces correspondingly smaller but verifiable changes. The key point is not magnitude but auditability: the observed feasibility shifts are attributable to concrete artefacts at pinned commits, demonstrating that the evaluator is sensitive to genuine engineering modifications and that its decision trace can be independently re inspected.

Overall, the fork based intervention provides a minimal but concrete validation that the framework responds to observable control implementation rather than to counterfactual score setting. It also illustrates a practical implication for remediation: improvements that target shared supply chain controls can incrementally move some techniques into the lowest feasibility band, but eliminating residual maximum feasibility cases requires addressing the missing core controls specific to those techniques.

\begin{table}[!t]
\caption{Pinned fork snapshots for concrete control implementation}
\label{tab:fork_manifest}
\centering
\footnotesize
\setlength{\tabcolsep}{4pt}
\renewcommand{\arraystretch}{1.15}
\begin{tabular}{p{1.6cm}p{1.55cm}p{1.55cm}p{1.6cm}}
\hline
Project & Baseline ID & Fork ID & Fork date \\
\hline
KServe & \texttt{5b033a4} & \texttt{e4ae8f7} & \texttt{2026-02-13} \\
vLLM   & \texttt{b17039b} & \texttt{de2222c} & \texttt{2026-02-13} \\
\hline
\end{tabular}
\end{table}

\begin{table}[!t]
\caption{Control score changes induced by concrete fork implementations (no manual score edits).}
\label{tab:fork_ctrl_delta}
\centering
\footnotesize
\setlength{\tabcolsep}{4pt}
\renewcommand{\arraystretch}{1.15}
\resizebox{\columnwidth}{!}{%
\begin{tabular}{lcc|cc|p{3.8cm}}
\hline
& \multicolumn{2}{c|}{KServe} & \multicolumn{2}{c|}{vLLM} & \\
Control ID & $s_b$ & $s_f$ & $s_b$ & $s_f$ & Observable fork evidence (paths/files) \\
\hline
\texttt{sc.sbom}                 & 0 & 2 & 0 & 2 & \texttt{.github/workflows/sbom.yml} \\
\texttt{sec.vulnerability\_scan} & 0 & 3 & 2 & 3 & \texttt{.github/workflows/codeql.yml} \\
% optional third row if you also implement hashing/signing evidence
% \texttt{model.hash\_release}      & 1 & <..> & 1 & <..> & \texttt{SHA256SUMS} / \texttt{*.sig} \\
\hline
\end{tabular}}
\end{table}

\begin{table}[!t]
\caption{Technique feasibility distributions before and after concrete fork implementations.}
\label{tab:fork_feasibility}
\centering
\footnotesize
\resizebox{\columnwidth}{!}{%
\begin{tabular}{lrrrr|rrrr}
\hline
& \multicolumn{4}{c|}{Baseline (upstream)} & \multicolumn{4}{c}{Fork (implemented)} \\
Project & $F=3$ & $F=2$ & $F=1$ & $F=0$ & $F=3$ & $F=2$ & $F=1$ & $F=0$ \\
\hline
KServe & 19 & 9 & 24 & 2 & 19 & 9 & 23 & 3 \\
vLLM   & 25 & 12 & 17 & 0 & 25 & 12 & 16 & 1 \\
\hline
\end{tabular}}
\end{table}

\subsection{Feasibility Shifts}

To explain how aggregate reductions arise, we inspect technique level feasibility distributions before and after unified hardening. Table~\ref{tab:feasibility_dist} reports, for each project, the counts of techniques at each discrete feasibility level $F \in \{0,1,2,3\}$. Since impact is fixed as $I_t^{(p)}=2$ for all techniques and projects, all distributional changes are driven solely by shifts in feasibility.

Across all projects, hardening increases the number of techniques in the lowest feasibility band, most visibly at $F=0$. KServe increases the count at $F=0$ from 2 to 9, and vLLM increases $F=0$ from 0 to 9. The same post hardening count of 9 techniques at $F=0$ is observed for Ultralytics, TorchServe, and Cassava Example. These movements provide an ordinal preserving view of improvement: they directly quantify how many techniques migrate to lower feasibility levels under the same consequence model.

The migration patterns differ by project and help interpret the earlier aggregate metrics. KServe exhibits no change in the highest feasibility bands, with $F=3$ and $F=2$ unchanged, while the reduction is concentrated in a shift from $F=1$ to $F=0$. This explains why the worst case contribution remains unchanged while the summary index decreases. By contrast, Ultralytics, TorchServe, and Cassava Example show visible reductions at $F=3$ alongside growth at $F=0$, indicating that hardening affects some techniques that previously satisfied the maximum feasibility predicate. For Cassava Example, $F=3$ decreases from 54 to 43, with remaining techniques redistributed primarily into $F=2$ and $F=0$.

These distributions also clarify why threshold based metrics can be insensitive to meaningful improvements. Under the thresholds $\tau_F=2$ and $\tau_I=2$ and with $I_t^{(p)}=2$ fixed, $HRS$ reduces to the fraction of techniques with $F \ge 2$. Improvements that move techniques within the region $F \in \{2,3\}$ do not change $HRS$, whereas transitions from $F=2$ to $F=1$ or from $F=3$ to $F=1$ do. Finally, the decrease in FISI is consistent with these downward feasibility shifts under a fixed policy, but remains a bounded heuristic summary rather than a calibrated estimate of real world risk magnitude.

\begin{table}[!t]
\caption{Technique feasibility Distributions Before and After Hardening}
\label{tab:feasibility_dist}
\centering
\footnotesize
\resizebox{\columnwidth}{!}{%
\begin{tabular}{lrrrr|rrrr}
\hline
& \multicolumn{4}{c|}{Baseline} & \multicolumn{4}{c}{Hardened} \\
Project & $F=3$ & $F=2$ & $F=1$ & $F=0$ & $F=3$ & $F=2$ & $F=1$ & $F=0$ \\
\hline
KServe & 19 & 9 & 24 & 2 & 19 & 9 & 17 & 9 \\
vLLM & 25 & 12 & 17 & 0 & 25 & 10 & 10 & 9 \\
Ultralytics & 42 & 11 & 1 & 0 & 38 & 7 & 0 & 9 \\
TorchServe & 30 & 20 & 4 & 0 & 25 & 20 & 0 & 9 \\
Cassava Example & 54 & 0 & 0 & 0 & 43 & 2 & 0 & 9 \\
\hline
\end{tabular}%
}
\end{table}

\section{Conclusion}

This paper proposed an evidence driven framework that operationalises AI security assessment as an executable decision function. Rather than relying on narrative justifications or assessor specific judgement, the framework anchors each assessment to externally inspectable artefacts and maps them into a stable, project independent control vocabulary, namely Control IDs. These controls are then evaluated by technique indexed rules aligned with MITRE ATLAS, enabling outputs that are repeatable across runs and comparable across projects under the same policy configuration.

A central contribution is the use of explicit evaluation semantics for how control evidence is composed into mitigation strength and how technique feasibility is selected under ordered predicates. The framework treats assessment as a structured computation over bounded ordinal inputs, with inspectable intermediate conditions and traceable outputs. By compiling ATLAS mitigation relations into an auditable rule base, it connects observable evidence to technique outcomes in a way that supports governance contexts where justification and replayability matter as much as the reported levels.

Because the framework is intended to function as a decision mechanism, we introduced formal verification to ensure that the implemented evaluator matches the declared semantics. We discharge bounded domain proof obligations for four properties. Totality ensures that every admissible input yields a defined feasibility output for each technique. Boundedness guarantees that mitigation strength values and feasibility levels remain within the declared ordinal range. Ordered semantic consistency ensures that the realised evaluator respects descending priority semantics and rules out priority inversions. Monotonicity guarantees that strengthening controls cannot increase assessed feasibility under the same semantics, supporting reliable hardening comparisons and preventing incentive inconsistent behaviour.

We evaluated the framework on five public open source AI related projects pinned to explicit snapshots, and studied two complementary intervention modes. Unified counterfactual hardening yields consistent reductions in aggregate summaries across all projects. Ordinal preserving analyses show feasibility migrating toward lower bands, while worst case contributions can persist when technique specific core controls remain missing under the observable evidence scope. To validate sensitivity to real engineering changes, we also implemented concrete controls in forked variants by adding SBOM generation and CI security gating as committed artefacts. Re assessment on pinned fork commits produces feasibility shifts attributable solely to newly observable evidence, demonstrating that the framework responds to auditable modifications rather than counterfactual score setting. We treat the Feasibility Impact Summary Index as a bounded heuristic under a fixed policy and report it as a secondary summary alongside ordinal preserving distributional and threshold metrics.

This study is intentionally evidence bound to repository visible artefacts to preserve reproducibility under pinned snapshots. As a result, it under approximates organisational control strength because many governance and operational controls are typically evidenced outside public repositories, including internal security reviews, supplier due diligence packages, runbooks, incident response exercises, and monitoring procedures. In our evaluation, such unobservable controls are conservatively treated as missing under the rubric, improving auditability but potentially overstating feasibility relative to a fully instrumented organisational assessment.

A second limitation is dependence on the instantiated policy configuration $\mathcal{P}$, including the mitigation to control mapping $\Phi$ and technique profiles, which encode governance choices rather than objective facts. We mitigate this by packaging $\mathcal{P}$ as a versioned, auditable configuration bundle and by reporting sensitivity results that expose when overly strict mappings saturate and lose discriminative power. A third limitation is that the primary outputs are bounded ordinal feasibility and impact levels, so any scalar aggregation should be interpreted as a heuristic summary rather than a calibrated estimate of real world risk magnitude. We therefore emphasise ordinal preserving metrics as primary evidence and use FISI only as a secondary aid under a fixed policy.

Future work will broaden evidence adapters beyond repository snapshots to incorporate verifiable enforcement artefacts and organisational evidence when available, enabling fuller control coverage without sacrificing auditability. We also plan larger scale evaluations across more projects and deployment settings, and systematic policy calibration procedures that preserve discriminative power while remaining conservative for the intended governance context.

\bibliographystyle{plainnat}
\bibliography{ref}

\appendix
\section{Control Scoring Rubric}\label{app:rubric}

\begingroup
\small
\setlength{\tabcolsep}{3pt}
\renewcommand{\arraystretch}{1.08}
\sloppy

\begin{longtable}{
  >{\raggedright\arraybackslash}p{3.2cm}
  >{\raggedright\arraybackslash}p{4.1cm}
  >{\raggedright\arraybackslash}p{\dimexpr\linewidth-8.0cm\relax}
}
\caption{Control ID scoring rubric (0 weakest, 3 strongest).}\label{tab:appendix_rubric}\\

\toprule
\textbf{Control ID} & \textbf{Meaning} & \textbf{Rubric anchors (0--3)} \\
\midrule
\endfirsthead

\caption[]{Control ID scoring rubric.}\\
\toprule
\textbf{Control ID} & \textbf{Meaning} & \textbf{Rubric anchors (0--3)} \\
\midrule
\endhead

\midrule
\multicolumn{3}{r}{continued on next page}\\
\endfoot

\bottomrule
\endlastfoot

\cid{api.abuse_detection} & Identify and suppress query based black box probes & \textbf{0}: None\newline \textbf{1}: Threshold alarm\newline \textbf{2}: Multi signal linkage\newline \textbf{3}: Automatic handling/isolation \\
\cid{api.access_control} & API level authentication and authorization & \textbf{0}: none\newline \textbf{1}: basic auth\newline \textbf{2}: RBAC/scopes\newline \textbf{3}: least privilege + MFA \\
\cid{api.exposure} & Public/Partner interface exposure boundaries and isolation & \textbf{0}: open\newline \textbf{1}: IP allowlist\newline \textbf{2}: tenant/network gated\newline \textbf{3}: policy+segmented \\
\cid{api.rate_limit} & Abuse/DoS Mitigation & \textbf{0}: no limit\newline \textbf{1}: static\newline \textbf{2}: per tenant\newline \textbf{3}: adaptive/dynamic \\
\cid{data.data_retention} & Minimum retention and scheduled cleanup & \textbf{0}: no policy\newline \textbf{1}: ad hoc\newline \textbf{2}: TTL configured\newline \textbf{3}: enforced \& verified \\
\cid{data.logging} & Minimization and traceability & \textbf{0}: raw\newline \textbf{1}: manual redaction\newline \textbf{2}: rules based\newline \textbf{3}: minimised \& centralised \\
\cid{data.user_notice} & Legality, transparency and options & \textbf{0}: none\newline \textbf{1}: notice only\newline \textbf{2}: consent/opt-out\newline \textbf{3}: audited consent \\
\cid{eval.bias_assessment} & Bias Identification and Mitigation & \textbf{0}: none\newline \textbf{1}: ad hoc\newline \textbf{2}: periodic tests\newline \textbf{3}: automated audits \\
\cid{eval.performance_monitoring} & Continuous performance monitoring & \textbf{0}: static\newline \textbf{1}: manual review\newline \textbf{2}: live KPIs\newline \textbf{3}: live + drift triggers \\
\cid{gpu.exclusive_process} & GPU exclusive processes prevent snooping/sampling & \textbf{0}: Shared\newline \textbf{1}: Attempting Exclusive Access\newline \textbf{2}: Stable Exclusive Access\newline \textbf{3}: Exclusive Access + Policy Drift Monitor \\
\cid{llm.corpus_audit} & Fine tuning/RAG corpus compliance & \textbf{0}: none\newline \textbf{1}: partial\newline \textbf{2}: documented\newline \textbf{3}: automated \& signed \\
\cid{llm.eval_backdoor} & Fine tuning backdoor detection & \textbf{0}: none\newline \textbf{1}: ad hoc\newline \textbf{2}: scheduled\newline \textbf{3}: gate in CI \\
\cid{llm.eval_coverage} & Evaluating Coverage and Regression & \textbf{0}: none\newline \textbf{1}: spot checks\newline \textbf{2}: planned suites\newline \textbf{3}: routine regression \\
\cid{llm.external_input}$\dagger$ & Whether to accept untrusted input & \textbf{0}: closed\newline \textbf{1}: limited\newline \textbf{2}: mixed\newline \textbf{3}: fully open \\
\cid{llm.fine_tuning} & Fine tuning processes and controls & \textbf{0}: uncontrolled\newline \textbf{1}: basic\newline \textbf{2}: templated\newline \textbf{3}: policy-gated \\
\cid{llm.ft_audit} & Fine-tuning governance and traceability & \textbf{0}: none\newline \textbf{1}: ad hoc\newline \textbf{2}: workflow\newline \textbf{3}: enforced \& audited \\
\cid{llm.input_filter} & Anti prompt injection/unsafe input & \textbf{0}: none\newline \textbf{1}: manual rules\newline \textbf{2}: hybrid/limited\newline \textbf{3}: continuous \& verified \\
\cid{llm.jailbreak_eval} & Security Red Team Assessment & \textbf{0}: absent\newline \textbf{1}: informal\newline \textbf{2}: structured\newline \textbf{3}: routine automated \\
\cid{llm.monitor} & Runtime monitoring & \textbf{0}: none\newline \textbf{1}: ad hoc\newline \textbf{2}: partial live\newline \textbf{3}: live with alerts \\
\cid{llm.output_filter} & Output side risk interception & \textbf{0}: none\newline \textbf{1}: keyword lists\newline \textbf{2}: rule+model\newline \textbf{3}: adaptive \& audited \\
\cid{llm.patch} & Quick patching/rollback & \textbf{0}: none\newline \textbf{1}: manual\newline \textbf{2}: partial automation\newline \textbf{3}: automated \& drilled \\
\cid{llm.plugin_review} & High risk capability gating & \textbf{0}: none\newline \textbf{1}: basic\newline \textbf{2}: structured\newline \textbf{3}: risk based gating \\
\cid{model.hash_release} & Version verification and traceability & \textbf{0}: none\newline \textbf{1}: hashes only\newline \textbf{2}: signed hashes\newline \textbf{3}: signed + registry provenance \\
\cid{org.access_control} & Key asset access control & \textbf{0}: none\newline \textbf{1}: basic\newline \textbf{2}: RBAC+MFA partial\newline \textbf{3}: least privilege + MFA \\
\cid{org.asset_inventory} & Clear scope of control and responsibilities & \textbf{0}: none\newline \textbf{1}: partial\newline \textbf{2}: complete\newline \textbf{3}: complete \& maintained \\
\cid{org.audit_trail} & Forensic traceability & \textbf{0}: none\newline \textbf{1}: fragmented\newline \textbf{2}: centralised\newline \textbf{3}: immutable \& \(\geq\)1y \\
\cid{org.drift_monitor} & Online quality protection & \textbf{0}: none\newline \textbf{1}: KPIs only\newline \textbf{2}: drift alarms\newline \textbf{3}: drift + retrain triggers \\
\cid{org.eol_disposal} & End of life cycle management & \textbf{0}: none\newline \textbf{1}: ad hoc\newline \textbf{2}: defined\newline \textbf{3}: enforced \& evidenced \\
\cid{org.eval_rerun} & Assess re-execution capability & \textbf{0}: none\newline \textbf{1}: ad hoc\newline \textbf{2}: scheduled\newline \textbf{3}: automated cadence \\
\cid{org.incident_response} & Emergency Preparedness & \textbf{0}: no plan\newline \textbf{1}: draft\newline \textbf{2}: trained\newline \textbf{3}: drilled \& improved \\
\cid{org.rollback} & Ability to roll back to a known good state & \textbf{0}: none\newline \textbf{1}: manual\newline \textbf{2}: partial auto\newline \textbf{3}: automated \& tested \\
\cid{org.threat_modeling} & Risk Assumptions and Coverage & \textbf{0}: none\newline \textbf{1}: ad hoc\newline \textbf{2}: documented\newline \textbf{3}: maintained \& reviewed \\
\cid{org.vdp} & External Disclosure and Response & \textbf{0}: none\newline \textbf{1}: policy only\newline \textbf{2}: active intake\newline \textbf{3}: triage \& SLAs \\
\cid{os.ptracetrap} & Blocking black box probes such as ptrace/perf\_event\_open & \textbf{0}: None\newline \textbf{1}: Partial\newline \textbf{2}: Comprehensive Strategy\newline \textbf{3}: Continuous Audit vs. Baseline Comparison \\
\cid{sc.model_provenance} & Model product authenticity/pedigree & \textbf{0}: untracked\newline \textbf{1}: stated\newline \textbf{2}: documented\newline \textbf{3}: cryptographically signed \\
\cid{sc.sbom} & Component transparency and supply chain visibility & \textbf{0}: absent\newline \textbf{1}: manual\newline \textbf{2}: automated\newline \textbf{3}: automated \& signed \\
\cid{sc.vendor_due_diligence} & External supplier/data source control & \textbf{0}: none\newline \textbf{1}: questionnaires\newline \textbf{2}: tracked\newline \textbf{3}: periodic re review \\
\cid{sc.version_pinning} & Preventing dependency drift and reproducibility & \textbf{0}: floating\newline \textbf{1}: partial\newline \textbf{2}: pinned/no verify\newline \textbf{3}: pinned \& verified \\
\cid{sc.weight_encryption} & Weight confidentiality protection & \textbf{0}: plaintext\newline \textbf{1}: ad hoc\newline \textbf{2}: at rest only\newline \textbf{3}: end to end + HSM \\
\cid{sec.cuda_tooling_restriction} & External probes and side channel inference reduction & \textbf{0}: None\newline \textbf{1}: Partial\newline \textbf{2}: Completely disabled or whitelisted\newline \textbf{3}: Continuous compliance scanning and drift alerts \\
\cid{sec.diagnostic_gating} & High observability/debuggability is subject to gating & \textbf{0}: No gating\newline \textbf{1}: Document exists but execution is not enforced\newline \textbf{2}: Role/environment gating\newline \textbf{3}: Signature + approval + audit closed loop \\
\cid{sec.penetration_test} & Independent defense verification & \textbf{0}: none\newline \textbf{1}: one off\newline \textbf{2}: periodic\newline \textbf{3}: annual + fixes verified \\
\cid{sec.redline_presence} & Redline upgrade configuration existence & \textbf{0}: none\newline \textbf{1}: draft\newline \textbf{2}: approved\newline \textbf{3}: enforced in tooling \\
\cid{sec.vulnerability_scan} & Vulnerability Management & \textbf{0}: never\newline \textbf{1}: ad hoc\newline \textbf{2}: scheduled\newline \textbf{3}: scheduled + blocked merges \\
\cid{vision.adv_train} & Enhancing robustness through training & \textbf{0}: none\newline \textbf{1}: pilots\newline \textbf{2}: partial coverage\newline \textbf{3}: broad \& maintained \\
\cid{vision.eval_coverage} & Evaluation system completeness & \textbf{0}: none\newline \textbf{1}: spot checks\newline \textbf{2}: planned suites\newline \textbf{3}: routine regression \\
\cid{vision.input_integrity} & Perception input validation and data integrity & \textbf{0}: none\newline \textbf{1}: manual/partial\newline \textbf{2}: automated limited\newline \textbf{3}: continuous \& verified \\
\cid{vision.input_randomization} & Raising the threshold for migration/query based attacks & \textbf{0}: None\newline \textbf{1}: Experimental\newline \textbf{2}: Deployed and passed regression testing\newline \textbf{3}: Enabled by default with downgrade protection \\
\cid{vision.label_audit} & Annotation quality management & \textbf{0}: none\newline \textbf{1}: ad hoc\newline \textbf{2}: sampled QA\newline \textbf{3}: systematic QA \\
\cid{vision.model_protect} & Model runtime hardening & \textbf{0}: none\newline \textbf{1}: basic\newline \textbf{2}: hardened\newline \textbf{3}: hardened \& attested \\
\cid{vision.monitor} & Runtime monitoring & \textbf{0}: none\newline \textbf{1}: ad hoc\newline \textbf{2}: partial live\newline \textbf{3}: live with alerts \\
\cid{vision.poison_scan} & Training data poisoning screening & \textbf{0}: none\newline \textbf{1}: ad hoc\newline \textbf{2}: automated checks\newline \textbf{3}: continuous + gating \\
\cid{vision.provenance} & Dataset provenance and lineage & \textbf{0}: untracked\newline \textbf{1}: stated\newline \textbf{2}: documented\newline \textbf{3}: signed \\
\cid{vision.redteam} & Adversarial evaluation/red teaming & \textbf{0}: none\newline \textbf{1}: informal\newline \textbf{2}: structured\newline \textbf{3}: automated \\
\cid{vision.robust_eval} & Robustness evaluation (evasion, patches, etc.) & \textbf{0}: none\newline \textbf{1}: ad hoc\newline \textbf{2}: scheduled suite\newline \textbf{3}: regression + gating \\
\cid{vision.training_audit} & Training pipeline traceability & \textbf{0}: none\newline \textbf{1}: partial\newline \textbf{2}: documented\newline \textbf{3}: automated \& signed \\

\end{longtable}

\endgroup

\end{document}